\documentclass{article}

\usepackage[preprint]{neurips_2025}

\usepackage[utf8]{inputenc} 
\usepackage[T1]{fontenc}    
\usepackage{hyperref}       
\usepackage{url}            
\usepackage{booktabs}       
\usepackage{amsfonts}       
\usepackage{nicefrac}       
\usepackage{microtype}      
\usepackage{multirow,multicol}
\usepackage{graphicx}
\usepackage{enumitem}
\usepackage{amsmath}
\usepackage[table]{xcolor}
\usepackage{wrapfig,lipsum,booktabs}
\usepackage{pifont}

\newcommand{\our}{Codifed Profiles\xspace}

\title{Codifying Character Logic in Role-Playing}

\author{
  Letian Peng, Jingbo Shang\thanks{$\ $Corresponding author.}\\
  Department of Computer Science\\
  University of California, San Diego\\
  \texttt{\{lepeng, jshang\}@ucsd.edu} \\
}

\usepackage{xspace}

\begin{document}

\maketitle

\begin{abstract}

This paper introduces \textbf{Codified Profiles} for role-playing, a novel approach that represents character logic as structured, executable functions for behavioral decision-making. Each profile defines a set of functions \texttt{parse\_by\_scene(scene)} that outputs a list of logic-grounded assertions \texttt{triggered\_statements}, using both explicit control structures (e.g., \texttt{if-then-else}) and condition checks like \texttt{check\_condition(scene, question)}, where each \texttt{question} is a semantically meaningful prompt about the scene (e.g., \textit{``Is the character in danger?''}) discriminated by the role-playing LLM as \textit{true}, \textit{false}, or \textit{unknown}.
This explicit representation offers three key advantages over traditional prompt-based profiles, which append character descriptions directly into text prompts:
(1) \emph{Persistence}, by enforcing complete and consistent execution of character logic, rather than relying on the model's implicit reasoning;  
(2) \emph{Updatability}, through systematic inspection and revision of behavioral logic, which is difficult to track or debug in prompt-only approaches;  
(3) \emph{Controllable Randomness}, by supporting stochastic behavior directly within the logic, enabling fine-grained variability that prompting alone struggles to achieve.  
To validate these advantages, we introduce a new benchmark constructed from 83 characters and 5{,}141 scenes curated from Fandom, using NLI-based scoring to compare character responses against ground-truth actions.  
Our experiments demonstrate the significant benefits of codified profiles in improving persistence, updatability, and behavioral diversity. 
Notably, by offloading a significant portion of reasoning to preprocessing, codified profiles enable even 1B-parameter models to perform high-quality role-playing, providing a scalable and efficient foundation for local deployment of role-play agents.
\footnote{Codes and datasets are available at \href{https://github.com/KomeijiForce/Codified_Profile_Koishiday_2025}{https://github.com/KomeijiForce/Codified\_Profile\_Koishiday\_2025}}

\end{abstract}

\begin{figure*}
    \centering
    \includegraphics[width=0.99\linewidth]{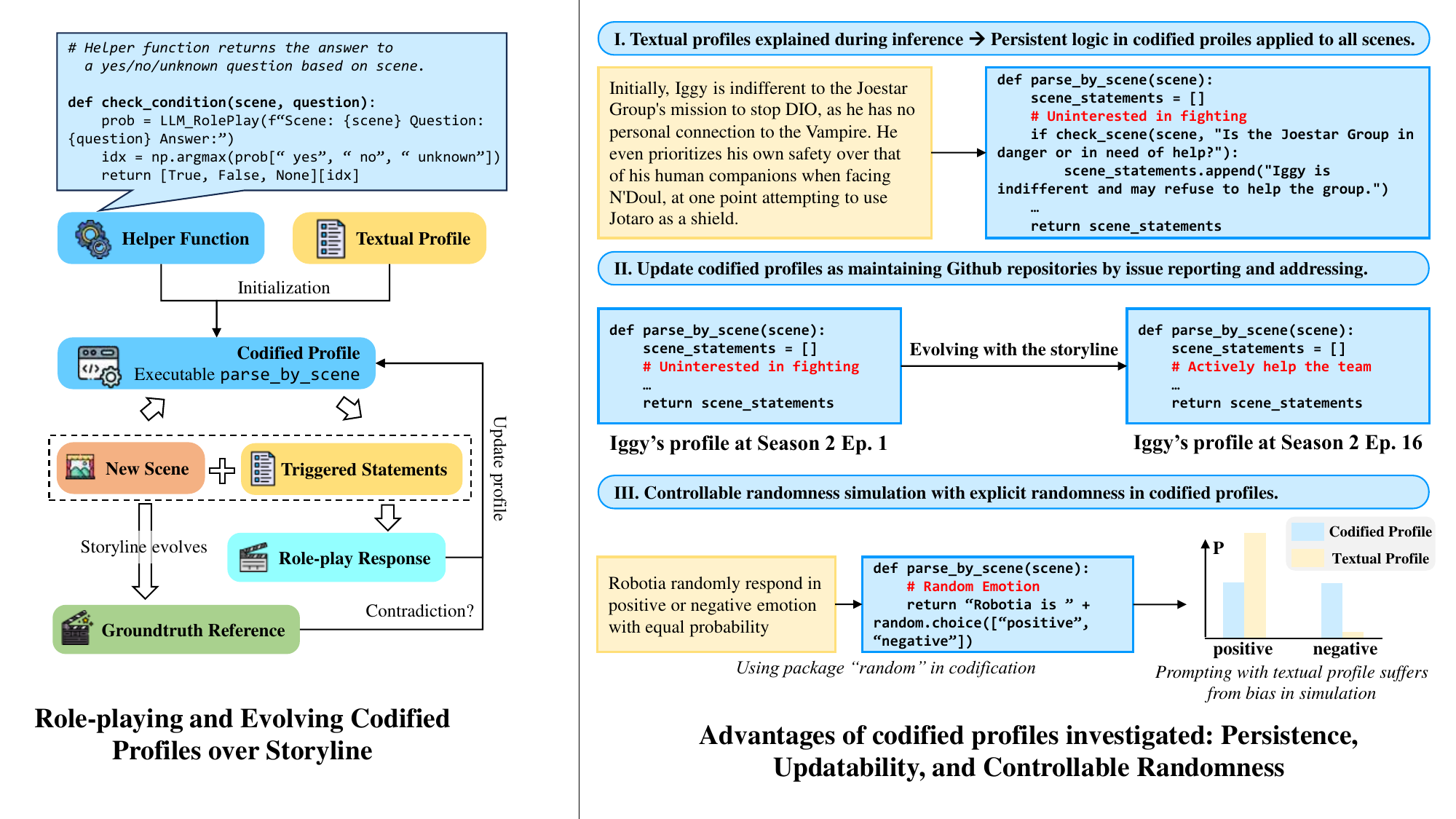}
    \caption{A presentation of the workflow and advantages of Codified Profiles.}
    \label{fig:ucp-intro}
\end{figure*}

\section{Introduction}

Role-playing~\citep{survey_rp,survey_rp_2} has emerged as an iconic capability of modern large language models (LLMs)~\citep{gpt-4, llama}, enabling them to simulate characters with distinct personas.
This ability has spurred a surge of interest in both research~\citep{eu,in-character,rolellm,mitigate_rp_hallucination} and applications\footnote{\href{https://character.ai/}{https://character.ai/}}, ranging from emotional companions and storytelling assistants~\citep{character_tell_story,CharacterMeet} to non-playable game characters~\citep{RPGBench} and metaverse agents~\citep{chatgpt-metaverse,survey_metaverse}.

Despite this progress, current LLM-based role-playing systems interpret character logic by appending character descriptions directly into prompts.
This approach \emph{lacks persistence and controllability}: character behavior is governed by the LLM's implicit reasoning, which is often brittle and inconsistent.  
Even well-crafted profiles frequently yield unpredictable results, particularly in complex or ambiguous scenes, and offer little transparency, making them \emph{difficult to debug or update}.  
In addition, interpreting profile logic at runtime imposes significant computational overhead and introduces latency, limiting scalability and responsiveness.

To address these challenges, we propose \textbf{Codified Profiles}, which compile character logic from natural language descriptions into executable functions.  
Unlike prompt-based role-playing, which appends character descriptions to prompts and relies on runtime interpretation, codified profiles are constructed in advance.  
They define character behavior using explicit control logic, producing consistent and interpretable outputs while reducing reliance on the LLM’s implicit reasoning.  
This design enhances both behavioral persistence and computational efficiency by eliminating the need for repeated, context-sensitive inference during interaction.

As illustrated in Figure~\ref{fig:ucp-intro}, we convert textual profile descriptions into \texttt{parse\_by\_scene(scene)} functions via LLM prompting.
Each function takes a scene as input, and based on explicit control structures such as \texttt{if-then-else}, returns a set of logic-grounded assertions (\texttt{triggered\_statements}) about what the character should do.  
Crucially, these functions include a semantic callable, \texttt{check\_condition(scene, question)}, which invokes the role-playing LLM to evaluate whether specific scene conditions hold.  
This callable returns \textit{true}, \textit{false}, or \textit{unknown}, determined by the logits over verbalizers (e.g., ``yes'', ``no'', ``unknown'').  
These checks allow codified profiles to go beyond symbolic matching, enabling flexible, semantic interpretation while preserving deterministic structure.  
At inference time, the codified function is applied to the current scene, and the resulting \texttt{triggered\_statements} guide the LLM’s response, grounding it in the character’s explicit logic.

Codified profiles offer three key advantages over prompt-based approaches.  
First, they provide \emph{persistence} by enforcing complete and consistent execution of character logic, rather than relying on the LLM’s implicit and often inconsistent reasoning.  
Second, they enable \emph{updatability}: since the logic is structured as executable code, developers, or even LLMs, can systematically inspect, debug, and revise specific behavioral rules over time, which is difficult to achieve with unstructured textual profiles.  
Third, they support \emph{controlled randomness}, allowing developers to explicitly encode stochastic behavioral branches within the logic itself.  
This provides fine-grained variability in character behavior that prompting alone struggles to produce reliably~\citep{bias_in_llm,probability_simulation}.  
Together, these properties make codified profiles especially well-suited for faithful, efficient, and adaptive role-playing, particularly in settings where smaller models are deployed or long-term consistency is required.

To evaluate the effectiveness of codified profiles, we construct a new benchmark comprising 83 characters and 5{,}141 scenes curated from Fandom, spanning mangas, novels, and television series.  
Each scene is paired with ground-truth character actions extracted from original narrative artifacts.  
To assess alignment between model outputs and expected behavior, we employ a validated LLM-based natural language inference (NLI) scoring framework~\citep{nli}, which measures logical entailment between generated responses and reference actions.

Our experiments start with a comparison between our codified profiles and traditional prompt-based profiles, and then dive deep into the effect of evolving profiles that update sequentially along the episode timeline, the influence of profile-driven randomness, and the model’s Best@K performance under stochastic response settings. 
Across all settings, codified profiles significantly improve behavioral consistency, adaptability, and diversity, especially when used with smaller LLMs.

We further investigate codification strategies, focusing on the granularity of profile segmentation.  
Our findings indicate that paragraph-level segments strike the best balance between fidelity and computational efficiency: larger segments often obscure fine-grained logic, while smaller ones lead to excessive fragmentation and redundant condition checking.  
We also analyze codification success rates and common pitfalls, such as misinterpreted conditions or invalid control structures, and present case studies that demonstrate the concrete advantages of codified profiles over prompt-based interpretations.

Our key contributions are as follows:
\begin{itemize}[nosep,leftmargin=*]
    \item We propose Codified Profiles, executable and interpretable representations of character logic that provide persistence, updatability, and controllable randomness beyond the limitations of prompt-based role-playing.
    \item We introduce a new benchmark, constructed from Fandom-sourced scenes and characters, for evaluating role-playing consistency. We plan to open-source this dataset to facilitate future research.
    \item We present an in-depth analysis of codification strategies, including optimal profile segmentation and codifiability metrics, and demonstrate their impact through detailed case studies.
\end{itemize}

\section{Background}

The rapid development of LLMs has not only raised the user's expectation of precise answers but also the ways that LLMs provide them, especially when asking for companionship and emotional support~\citep{survey_rp,survey_rp_2}. Role-playing emerges as a newborn and trending task in accordance with such demand, which enables preferable interactions with LLMs~\citep{conversation_qa_task,lamda}, for instance, simulating interactions with fictional characters. Role-playing also allows LLMs to simulate different individuals taking different profiles, serving as an engine for storytelling~\citep{character_tell_story,CharacterMeet}, games~\citep{RPGBench}, and even metaverse~\citep{chatgpt-metaverse,survey_metaverse}. We provide a full overview of current progress in role-playing, covering training, inference, grounding, and evaluation. We also discuss the role of codes in role-playing in traditional games and modern LLMs.

\paragraph{Training and Inference} The most straightforward way to build a character into an LLM is to train on how the character reacts to different conditions. Such training data can be obtained from existing plots~\citep{chatharuhi} or profile-based data synthesis~\citep{eu}. However, the scarcity of plots (especially for original characters) and insufficient grounding for synthesis~\citep{apc} cause hallucination challenges~\citep{mitigate_rp_hallucination} caused by fine-tuning. During the inference, LLMs are generally prompted with statements from profiles to mitigate hallucinations, which requires a strong grounding system.

\paragraph{Grounding and Updating} The grounding system in role-playing models what and how to append profile information to prompts. The ``how'' part is modeled similarly to traditional retrieval-augmented generation (RAG)~\citep{rag} using similarity metrics to retrieve the most relevant statements~\citep{dpr,recap,lamp}. The ``what'' part discusses how to preprocess the statements for retrieval, including formatting and segmentation~\citep{character_profiling}. Advanced role-playing systems also expect statements to be dynamically updated during the interaction~\citep{memory_mechanism_personalization} for both working memory and long-term profile. Such updates generally ``backpropagate'' self-reflection~\citep{self-reflection} to the profile, which refines the original statements in a textual gradient style~\citep{textgrad}.

\paragraph{Evaluation} There are evaluations for both understanding and generation in role-playing. For understanding, factuality checking and multiple-choice question answering benchmarks~\citep{lamp,character_profiling} are developed to evaluate the understanding of characters. For generation, the reference comes from ground-truth scenes~\citep{rolellm} and human judgment~\citep{hpd,real_world_long_term}. The profile is an essential reference for human evaluators, especially when evaluators are unfamiliar with the character. The evaluation can also be decomposed into simple tasks like natural language inference~\citep{nli} that can be trustfully assigned to neural models for automated evaluation~\citep{apc}. Some works also apply LLMs to evaluate more advanced attributes such as value and personality~\citep{eu}.

\paragraph{Code in Role-playing} Building characters with code is strongly connected to the development of video games~\citep{nero,llm_game}. The most common practice is to hardcode the actions of non-player characters, making them actors in prewritten scripts. They are only allowed to intellectually interact with players in some subscenes (e.g., battle, chess), which hardly affect the storyline progression. In role-playing, codes and LLMs represent two extremes in interaction freedom, which are synergized in our \our framework to reach a balanced point by generating and executing prewritten symbolic logic by a flexible neural system.

\section{Codified Profile}

\subsection{Role-playing Preliminary}

A pure role-playing system models character logic as ``Input a scene $s$, output a response $r$ to the scene as character X'', where the response can be utterances, actions, or both. Here we use $\text{LLM}(\cdot)$ to represent the role-playing systems as our discussion scope will be LLM-driven role-playing. The LLM is prompted to role-play as characters with a system prompt $I_\text{role-play}$, which does not contain information of any character to be universally applied for role-playing.

\begin{equation}
\small
    r = \text{LLM}(s\mid I_\text{role-play})
\end{equation}

Textual profile $P=[p_1, p_2, \cdots, p_N]$ ($p_i$ represents segments of $P$ like paragraphs) further grounds the role-playing with in-context information about character logic. With $P$ appended to the input, the role-playing LLM can output its response $r$ directly or after reasoning with chain-of-thoughts (CoT)~\citep{chain_of_thoughts} $t$ prompted from CoT reasoning instruction $I_\text{CoT}$.

\begin{equation}
\small
    r = \text{LLM}(s\mid P, t, I_\text{role-play}, I_\text{CoT})\ \textrm{where}\ t = \text{LLM}(s\mid P, I_\text{role-play}, I_\text{CoT})
\end{equation}

\subsection{Codifying Profiles}

Our motivation to codify profiles emerges from the instability for LLMs to interpret textual profiles during inference (the CoT stage $t = \text{LLM}(s\mid I_\text{role-play}, P)$), which completely relies on LLM's reasoning accuracy without missing or misinterpretation. This reliance creates a brittle dependency on the model’s reasoning over long, natural language inputs, often leading to errors in complex scenes. Codification mitigates this fragility by offloading as much reasoning as possible into deterministic control logic, allowing the LLM to focus on localized decisions such as condition evaluation and action selection.\footnote{All prompts and templates for implementation are appended in Appendix~\ref{apdx:prompt_template}.}

In practice, we utilize LLM's coding ability to codify each profile segment $p_i$ (e.g., segmented by paragraph) into an executable function $f_i: s \rightarrow p^\text{trig}_i$ (Code implementation: \texttt{parse\_by\_scene(scene)} $\rightarrow$ \texttt{triggered\_statements}), which returns the possible reactions $p^\text{trig}_i$ in the scene $s$ based on the logic written in $p_i$. LLM then combines all $p^\text{trig}_1$ to summarize a final response $r$.

\begin{equation}
\small
    r = \text{LLM}(s\mid [f_1(s), f_2(s), \cdots, f_N(s)], I_\text{role-play})
\end{equation}

To enhance flexibility while preserving structure, each function $f_i$ can incorporate semantic checks using a callable helper function \texttt{check\_condition(scene, question)}.
This callable allows $f_i$ to evaluate context-sensitive conditions that are difficult to express through static rules alone.
For example, a rule like ``Whenever being insulted, X will become outrageous'' may be codified as:

\begin{center}
\small
\texttt{if check\_condition(scene, "Is X being insulted?"):}\\
    \quad \quad \quad \quad \quad \texttt{triggered\_statements.append("X become outrageous")}
\end{center}


Internally, \texttt{check\_condition} queries the role-playing LLM with a natural language question and interprets the logits over verbalizers (e.g., \textit{``yes'', ``no'', ``unknown''}) representing boolean (True/False) and uncertain (None) judgments.
This mechanism allows codified profiles to blend deterministic control flow with semantically rich interpretation, enabling nuanced yet persistent character behavior across diverse scene inputs. For further clarity, we include several codification cases in Figure~\ref{fig:extended_case_codification} of Appendix~\ref{apdx:extended_case} about personality, relation, and working mechanism of superpower. As condition checking is easier than full-scene reasoning, codification shifts the burden from generation to classification, allowing smaller LLMs to perform competitively by focusing on localized judgments. This makes codified profiles especially effective for low-resource settings, where smaller models can still produce coherent and contextually grounded behavior.

\subsection{Evolving by Storyline}
\label{sec:evolving}

\begin{figure*}
    \centering
    \includegraphics[width=0.99\linewidth]{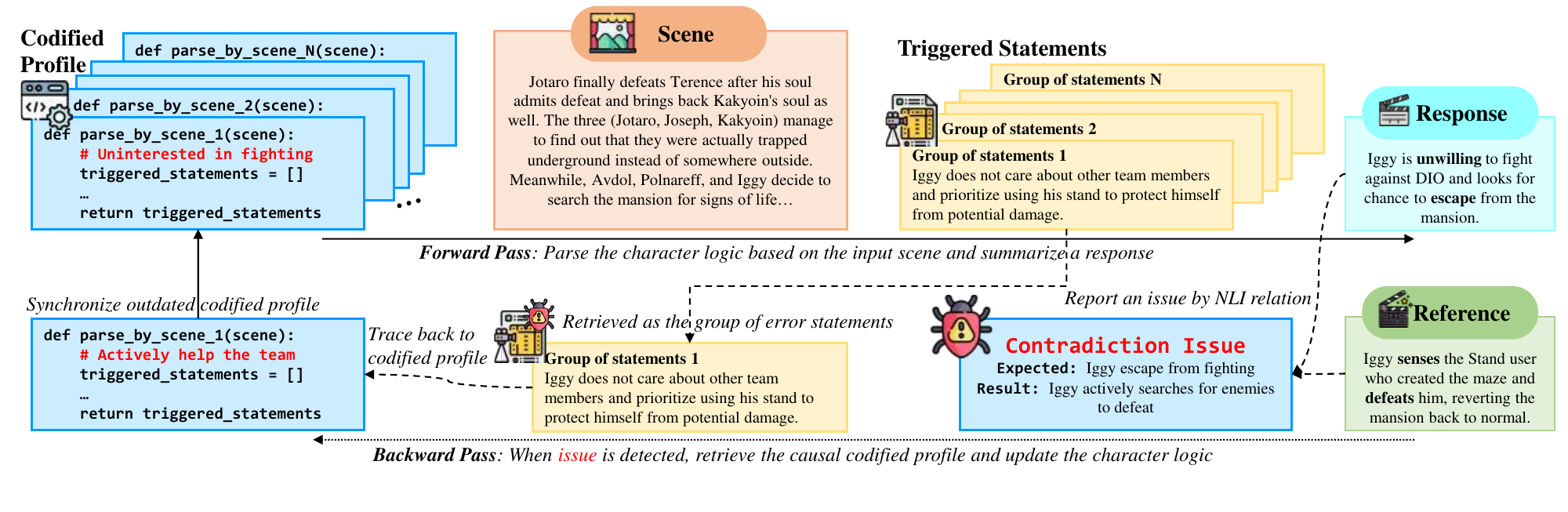}
    \vspace{-5mm}
    \caption{The evolving mechanism of codified profile to synchronize character with the storyline.}
    \vspace{-3mm}
    \label{fig:evolving}
\end{figure*}

An essential advantage of \our is long-term maintenance, considering the Github repository updates according to discovered issues. Codified profiles can also be updated along with the storyline, and each profile version reflects the character logic in different episodes. We plot the workflow to evolve codified profile in Figure~\ref{fig:evolving}. Issues in role-playing emerge from predicted response $r$ mismatching with ground-truth response $r_{\text{ref}}$ based on scene $s$. We select natural language inference (NLI) as the initial signal for profile updating, which discriminates $r$ as \{\textit{``contradicted''}, \textit{``neutral''}, \textit{``entailed''}\} based on $r_{\text{ref}}$.

When $\text{NLI}(r_\text{ref}, r) \in \{\textit{``contradicted''}, \textit{``neutral''}\}$, the issue will be reported to the response generation stage, citing \textit{``contradicted statement''} (contradicted) or \textit{``relevant but not detailed statement''} (neutral). Then, LLM will be prompted to detect the group of \texttt{triggered\_statements} that fits such accusation the most. Supposing $p^\text{trig}_j$ to be the proposed \texttt{triggered\_statements}, $f_j$ will consequently be revised as a problematic character logic. Finally, $f_j$ will be updated based on the input scene $s$ together with $p^\text{trig}_j$, expected response $r_\text{ref}$, and result response $r$ to synchronize the character logic with the storyline.

\subsection{Controllable Randomness Simulation}

LLMs are known to fail to simulate precise randomness because their probability is controlled by temperature in generation, which makes it almost impossible to calibrate them to precisely simulate multiple random behaviors. For instance, when prompted to respond \textit{``positively or negatively in equal probability''}, LLMs will be biased to one of the sentiments due to their training data distribution or decoding preferences, leading to inconsistent and unbalanced behavior~\citep{bias_in_llm,probability_simulation}.

Codified profile addresses this limitation by externalizing randomness into explicit control logic. Rather than relying on the LLM to probabilistically choose responses at generation time, we prompt it to generate executable Python code that encodes randomness via constructs such as \texttt{random.choice([...])} and \texttt{random.random() < p}. This allows us to define and enforce precise stochastic behavior within the profile logic itself. For example, to simulate the whimsy robot:

\begin{center}
\small
\texttt{sentiment\_statement = random.choice(["X is positive.", "X is negative."])} \\
\texttt{triggered\_statements.append(sentiment\_statement)} \quad \quad \quad \quad \quad \quad \quad \quad \quad \quad \quad \quad \quad \quad \quad
\end{center}

By embedding such logic in codified profiles, we achieve fine-grained control over behavioral randomness, ensuring both reproducibility and tunability. This approach is particularly advantageous in applications where controlled variability is essential such as user interaction.

\section{Fandom Benchmark}

While several benchmarks have been developed for evaluating role-playing capabilities of LLMs, most of them primarily focus on dialogue performance between the character and a user, rather than assessing the model's ability to enact complex, situational behaviors within grounded scenes. Moreover, existing benchmarks often rely on data synthesized by LLMs or sparsely annotated scripts~\citep{rolellm,survey_rp}, which restricts them in evaluating an in-depth understanding of massive character logic.

\begin{figure*}
    \centering
    \includegraphics[width=0.99\linewidth]{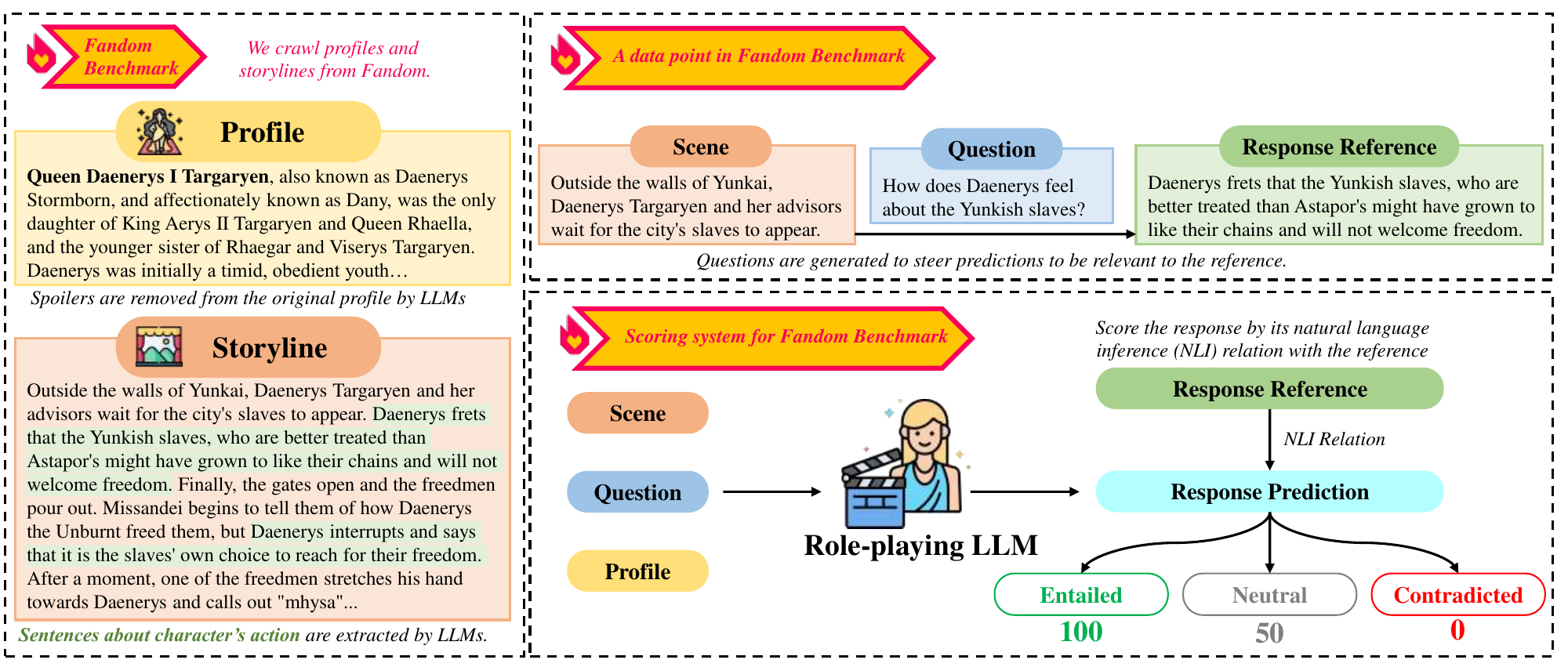}
    \vspace{-3mm}
    \caption{Curation and evaluation scenarios of our Fandom Benchmark.}
    \label{fig:fandom-benchmark}
\end{figure*}

To address these limitations, we propose using Fandom~\footnote{\href{https://www.fandom.com/}{https://www.fandom.com/}} as a data source for building a behavior-centric role-playing benchmark. Fandom offers extensive, high-quality character profiles and structured storyline summaries across a wide range of fictional universes. We directly utilize these summaries to construct our benchmark: for each episode or narrative segment, we prompt an LLM in an extractive manner to identify sentences that describe a character’s concrete actions, and treat the preceding text as the corresponding scene context. To prevent the role-playing LLM's responses from drifting beyond the scope of evaluation, we additionally generate a guiding question for each scene, carefully designed to constrain the response without providing clues toward the correct answer.

During evaluation, the role-playing LLM is given the character’s profile and the scene, and tasked with predicting the character’s next action based on the question. To avoid information leakage, profiles are filtered to remove spoilers. Predicted actions are then compared to ground-truth actions using an LLM-based NLI scoring system, where predictions that are entailed by the ground truth receive a score of 100, neutral responses 50, and contradicted ones 0. 

We select $6$ reputational artifacts to cover various themes of stories. 

\begin{itemize}[nosep,leftmargin=*]
    \item \textbf{Haruhi}: \textit{``Suzumiya Haruhi''} is a high school sci-fi comedy centered on a girl who unknowingly possesses godlike powers, dragging her eccentric club into supernatural chaos.
    \item \textbf{K-On!}: \textit{``K-On!''} follows a group of high school girls who form a light music club, bonding over tea, friendship, and their shared love for music.
    \item \textbf{JOJO}: \textit{``JOJO's Bizarre Adventure''} is a multi-generational saga where members of the Joestar family battle supernatural threats with flamboyant powers and style. We select season $3$, \textit{``Stardust Crusaders''}, which describes Jotaro's bizarre advantage from Japan to Egypt, fighting against DIO.
    \item \textbf{FMA}: \textit{``Full Metal Alchemist''} tells the story of two brothers who use alchemy in a perilous quest to restore their bodies after a tragic experiment. We select the 2009 edition of FMA.
    \item \textbf{AGOT}: \textit{``A Game of Thrones''} is a gritty political fantasy where noble families vie for power in a brutal world of betrayal, war, and dragons. We focus on the first $3$ seasons of the TV series.
    \item \textbf{ATLA}: \textit{``Avatar: The Last Airbender''} is an epic tale of a young boy who must master all four elements to stop a tyrannical war and restore the world's balance. We focus on \textit{``Book One: Water''}.
\end{itemize}

We select $83$ characters from these artifacts and extract $5,141$ scenes with their participation, which offers a comprehensive benchmarking of different role-playing strategies. We list the characters in our experiments in Appendix~\ref{apdx:character_info} together with brief introductions for readers unfamiliar with the involved artifacts. An example profile and testing cases are also appended in this Appendix~\ref{apdx:character_info} for better benchmark clarity.

\begin{table}
\centering
\small
\scalebox{.88}{
\begin{tabular}{lccccccccc}
\toprule
{Statistics (Main/Minor)} & {Haruhi} & {K-On!} & {JOJO} & {FMA} & {AGOT} & {ATLA} \\
\midrule
\#Character & $5$/- & $5$/$4$ & $7$/$9$ & $5$/$7$ & $11$/$19$ & $4$/$7$ \\
\#Scene & $48.4$/- & $173.0$/$49.8$ & $111.6$/$20.8$ & $85.8$/$32.9$ & $63.0$/$23.1$ & $214.5$/$31.1$ \\
\#Word per profile & $782.6$/- & $676.4$/$449.2$ & $925.0$/$488.8$ & $1564.6$/$1351.1$ & $1517.3$/$1243.5$ & $1578.2$/$740.1$ \\
\#Paragraph per profile & $11.8$/- & $9.2$/$7.0$ & $15.7$/$9.9$ & $16.0$/$15.1$ & $14.6$/$11.8$ & $14.5$/$9.3$ \\
\bottomrule
\end{tabular}
}
\vspace{2mm}
\caption{Important statistics of the Fandom Benchmark.}
\label{tab:stats}
\end{table}

\paragraph{Statistics} Statistics in Table~\ref{tab:stats} highlight the depth and breadth of the Fandom Benchmark. The average number of scenes for main characters ranges from 48 to 214 scenes each, with minor characters adding dozens more, resulting in a total of 5,141 annotated scenes. Profiles are notably rich, averaging around 1{,}000 words and 15 paragraphs for main characters, with similarly detailed coverage for minors. This combination of long-form profiles and large-scale scene coverage offers a rigorous testbed for evaluating models' ability to reason over sustained character logic, adapt behavior across contexts, and handle diverse narrative situations with consistency.

\section{Experiment}

\subsection{Evaluation Scenarios}

\begin{wraptable}{r}{0.5\textwidth}
  \vspace{-3mm}
  \begin{center}
  \small
    \scalebox{0.78}{\begin{tabular}{lp{1.0cm}p{1.2cm}p{0.8cm}p{2.2cm}}
\toprule
Experiment & Update Profile & Import Random & Temp. & Metric\\
\midrule
Basic & \ding{55} & \ding{55} & \ding{55} & NLI\\
Evolving & \ding{51} & \ding{55} & \ding{55} & NLI\\
Stochastic & \ding{55} & \ding{51} & \ding{51} & Best@K (NLI)\\
\bottomrule
\end{tabular}}
  \end{center}
  \vspace{-3mm}
  \caption{Detailed setups in experiments.}
  \label{tab:exp_setup}
  \vspace{-3mm}
\end{wraptable}

For evaluation, we adopt a two-stage setup involving separate LLMs for codification and role-playing. First, we use \texttt{gpt-4.1} as the codification LLM to convert natural language character profiles into codified profiles. Then, we employ \texttt{llama-3.1-8b-instruct} as the role-playing LLM. This model is chosen to ensure experimental efficiency, including tractable forward pass counts and full access to output logits for analysis.\footnote{We also find gpt-4 series memorizing scenes in famous artifacts like \textit{``A Game of Thrones''}, injecting bias into evaluation.} For each scene in our Fandom benchmark, the role-playing LLM is prompted with a given profile and is asked to produce the character’s next action. This response is then scored using LLM-based (\texttt{gpt-4.1}) NLI against the ground-truth action, as mentioned in the benchmark section. We further extend the basic test scenario to evolving profile and stochastic response, validating the advantage of codified profiles in character logic synchronization and stochastic response. Difference between experiment setups can be referred to Table~\ref{tab:exp_setup}.

\paragraph{Evolving Profile} In this setting, scenes are arranged in temporal order, and evaluation proceeds in a loop of test on $i$-th scene, evolve (if not entailed), then test on $(i+1)$-th scene to efficiently benchmark adaptation over time. For each scene, the predicted response is compared with the reference using NLI; if the result is contradicted or neutral, the profile is updated to reflect new behavioral logic, as described in $\S$~\ref{sec:evolving} our methodology section.

\paragraph{Stochastic Response} Here, the role-playing LLM generates multiple responses per scene using a fixed codified profile that encodes probabilistic behavior (e.g., \texttt{random.choice}) and generation temperature (set to under $0.7$ for quality). Codification LLM is explicitly prompted to include randomness into the control flow. We assess performance using Best@K scoring, which selects the highest NLI score among K sampled outputs, reflecting the expected alignment with the reference behavior across multiple samplings.

\subsection{Baselines}

We include various baselines and role-playing mechanisms to validate our claims.

\begin{itemize}[nosep,leftmargin=*]
    \item \textbf{Vanilla} The role-playing LLM is prompted with only the scene. This setting serves as a lower bound, representing performance with parameterized knowledge when profile data is unavailable.
    \item \textbf{Textual Profile} The full natural language character profile is provided alongside the scene. This is the standard grounding method in most existing role-playing systems.
    \item \textbf{Codified RAG} A simple codification strategy that implements retrieval-augmented generation (RAG) over the profile. Each segment of the profile is wrapped in a \texttt{check\_condition} function to determine its relevance to the current scene (\texttt{if check\_condition(scene, question): triggered\_statements.append(segment)}). This method evaluates whether logic-based codification offers deeper character understanding than shallow relevance matching.
    \item \textbf{Reasoning Mechanism} We incorporate chain-of-thoughts prompting before the action prediction step. To fairly evaluate reasoning efficiency, we apply CoT both on top of the Textual Profile and the Codified Profile, and vary the average reasoning chain length from 0 upward. For each configuration, we report both performance and average forward pass count to assess the trade-off between reasoning quality and computational cost.
\end{itemize}

\subsection{Role-playing Results}

\begin{table}
\centering
\small
\scalebox{.99}{
\begin{tabular}{llcccccccc}
\toprule
\multicolumn{2}{l}{{Artifact}} & {Haruhi} & {K-On!} & {JOJO} & {FMA} & {AGOT} & {ATLA} & Average \\
\midrule
\multirow{7}*{\rotatebox{90}{Main}}& \cellcolor{gray!20} \#Character & \cellcolor{gray!20}$5$ & \cellcolor{gray!20}$5$ & \cellcolor{gray!20}$7$ & \cellcolor{gray!20}$5$ & \cellcolor{gray!20}$11$ &  \cellcolor{gray!20}$4$ & \cellcolor{gray!20}$5.3$ \\
\cmidrule(lr){2-9}
& Vanilla (No Profile) & $63.53$ & $63.60$ & $59.70$ & $67.10$ & $64.93$ & $66.40$ & $64.21$ \\
& Original Textual Profile & $70.53$ & $68.47$ & $60.64$ & $68.02$ & $61.52$ & $66.70$ & $65.98$ \\
& Codified RAG & $69.03$ & $66.81$ & $59.52$ & $69.14$ & $65.39$ & $66.33$ & $66.04$ \\
& Codified Profile & $72.14$ & $69.23$ & $63.92$ & $69.60$ & $67.51$ & $67.90$ & $\textbf{68.38}$ \\
\cmidrule(lr){3-9}
& $\quad$ + Textual Profile & $72.81$ & $71.17$ & $63.42$ & $71.36$ & $65.60$ & $67.37$ & $\textbf{68.62}$ \\
\midrule
\multirow{7}*{\rotatebox{90}{Minor}}& \cellcolor{gray!20} \#Character & \cellcolor{gray!20} & \cellcolor{gray!20}$4$ & \cellcolor{gray!20}$9$ & \cellcolor{gray!20}$7$ & \cellcolor{gray!20}$19$ & \cellcolor{gray!20}$7$ & \cellcolor{gray!20}$9.2$ \\
\cmidrule(lr){2-9}
& Vanilla (No Profile) & & $65.12$ & $62.53$ & $61.17$ & $67.65$ & $66.89$ & $64.67$ \\
& Original Textual Profile & & $65.88$ & $64.70$ & $65.83$ & $66.21$ & $65.88$ & $65.70$ \\
& Codified RAG & & $63.28$ & $68.13$ & $66.97$ & $68.62$ & $66.65$ & $66.73$ \\
& Codified Profile & & $68.59$ & $70.06$ & $68.21$ & $70.51$ & $71.95$ & $\textbf{69.87}$ \\
\cmidrule(lr){4-9}
& $\quad$ + Textual Profile & & $67.97$ & $68.94$ & $69.22$ & $71.83$ & $71.86$ & $\textbf{69.96}$ \\
\bottomrule
\end{tabular}
}
\vspace{2mm}
\caption{Role-playing performance based on different profiles.
}
\label{tab:main}
\end{table}

\begin{wrapfigure}{r}{0.5\textwidth}
  \begin{center}
    \includegraphics[width=0.5\textwidth]{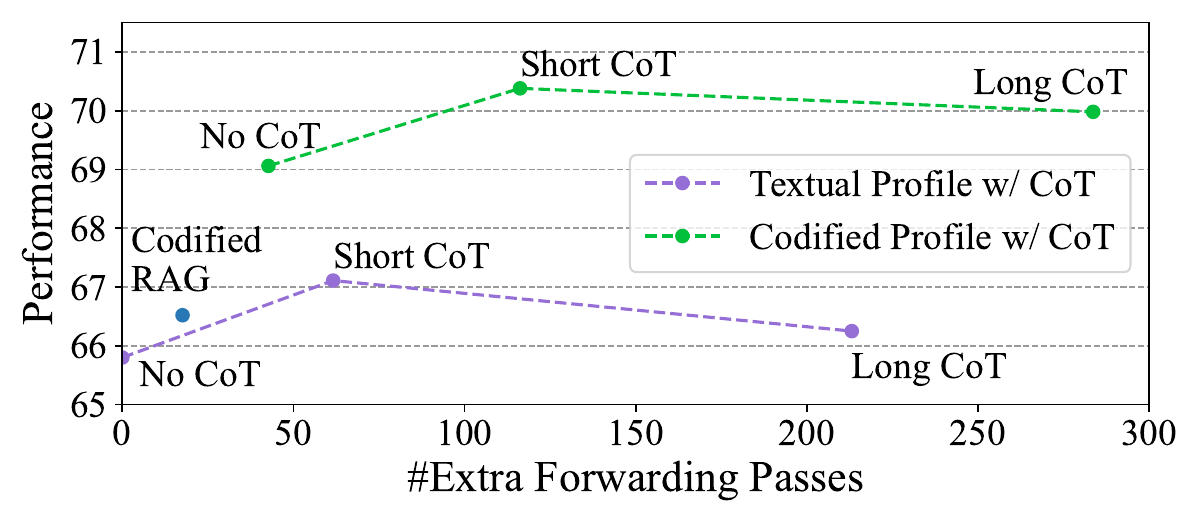}
  \end{center}
  \vspace{-5mm}
  \caption{The role-playing performance with reasoning mechanism.}
  \vspace{-5mm}
  \label{fig:ecp_reasoning}
\end{wrapfigure}

In Table~\ref{tab:main} and Figure~\ref{fig:ecp_reasoning}, we plot the performance of different role-playing methods. The result first demonstrates the effectiveness of grounding in role-playing, as all grounding-based methods outperform the vanilla (No Profile) baseline, highlighting the necessity of contextual character guidance. This supports our central claim that enforcing consistent and complete execution of character logic yields better results than relying on the LLM’s implicit and often inconsistent reasoning, as Codified Profiles outperform Original Textual Profiles across both main and minor characters. Furthermore, Codified Profiles also surpass Codified RAG, showing that their strength lies not merely in retrieving relevant information but in expressing structured, executable behavior logic that leads to more persistent and faithful character simulation.

Adding textual profiles on top of codified profiles provides further but marginal improvements, especially when the performance difference between code and text is small. This suggests that textual prompts can sometimes complement information that may have been simplified or abstracted during codification, enhancing the expressiveness of the final behavior.

\paragraph{Reasoning} Figure~\ref{fig:ecp_reasoning} compares codified reasoning with chain-of-thoughts reasoning, averaged over all characters. The results show that the codified reasoning flow achieves strong performance with fewer reasoning steps than chain-of-thoughts, demonstrating it as a more efficient grounding mechanism compared to approaches that rely heavily on runtime reasoning. Notably, performance improves further when codified profiles are combined with targeted reasoning, indicating that our method can benefit from additional inference when applied judiciously. In contrast, the long chain-of-thought (CoT) baseline underperforms, and manual inspection reveals that it often suffers from overthinking, introducing unnecessary complications that derail character-consistent behavior.

\begin{figure}[ht]
    \centering
    \includegraphics[width=1.0\linewidth]{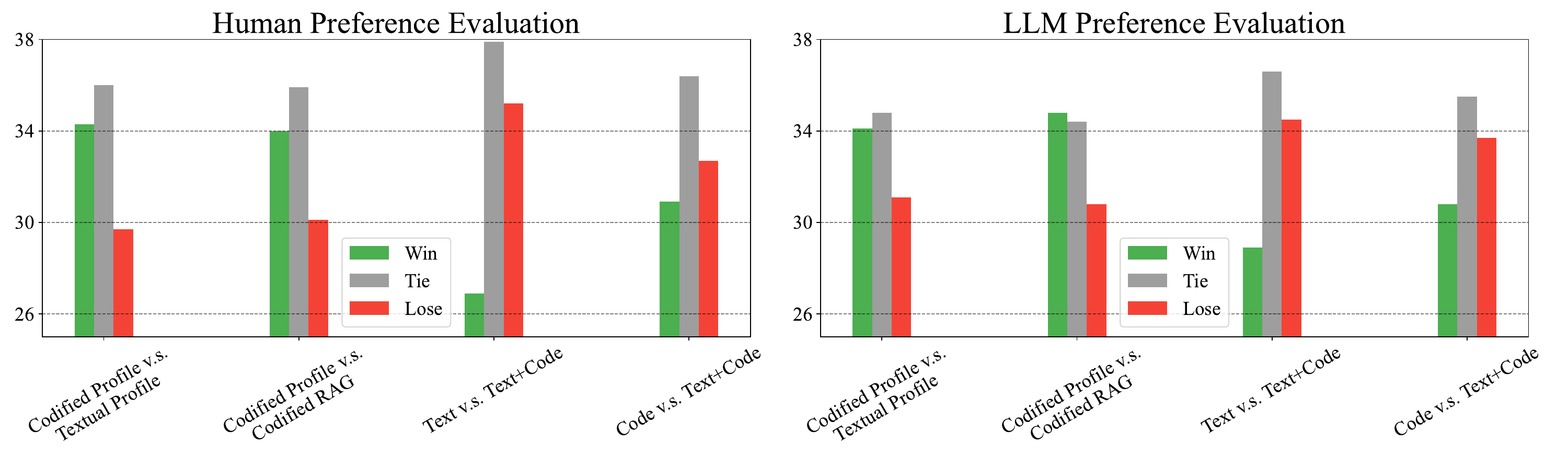}
  \vspace{-5mm}
    \caption{The Human/LLM-based preference evaluation results.}
    \label{fig:ecp_preference}
\end{figure}

\paragraph{Human/LLM Preference Evaluation} A limitation of the NLI score is its inability to capture fine-grained differences in response quality, since responses with the same NLI relation (e.g., entailed) are assigned the same score. To address this, we conduct a preference-based evaluation where both humans and LLMs are asked to choose the better response between two candidates, given the reference. LLMs are prompted on all scenes with responses shown in both orders to mitigate position bias, and a tie is recorded when the model selects different responses in two runs. Human evaluators assess 5 scenes per character, selecting a preferred response or indicating a tie.

Figure~\ref{fig:ecp_preference} presents the results of this evaluation. First, codified profiles are preferred over both textual profiles and Codified RAG by both human and LLM judges, indicating their advantage in producing precise and contextually faithful responses. While Codified RAG achieves higher NLI scores than textual profiles, it loses the advantage in preference evaluations, likely due to truncation or omission of relevant details during retrieval. Second, ensembling codified and textual responses further improves performance, though codified-only responses contribute more wins, suggesting they provide the stronger baseline. Notably, we observe that LLM-based preferences tend to result in more ties and occasional misjudgments, likely due to distraction by irrelevant details, reinforcing the importance of human judgment in evaluating nuanced role-play quality.

In Appendix~\ref{apdx:character_comparison}, We further break down the evaluation by character and observe that codified profiles are favored for logic-heavy ``fickle'' characters (e.g. Littlefinger in AGOT), while textual profiles perform better for emotionally expressive characters (e.g. Winry in FMA). This complementarity helps explain the ensemble’s occasionally improved performance by combining structured reasoning with naturalistic expression.

\begin{figure}[ht]
    \centering
    \includegraphics[width=1.0\linewidth]{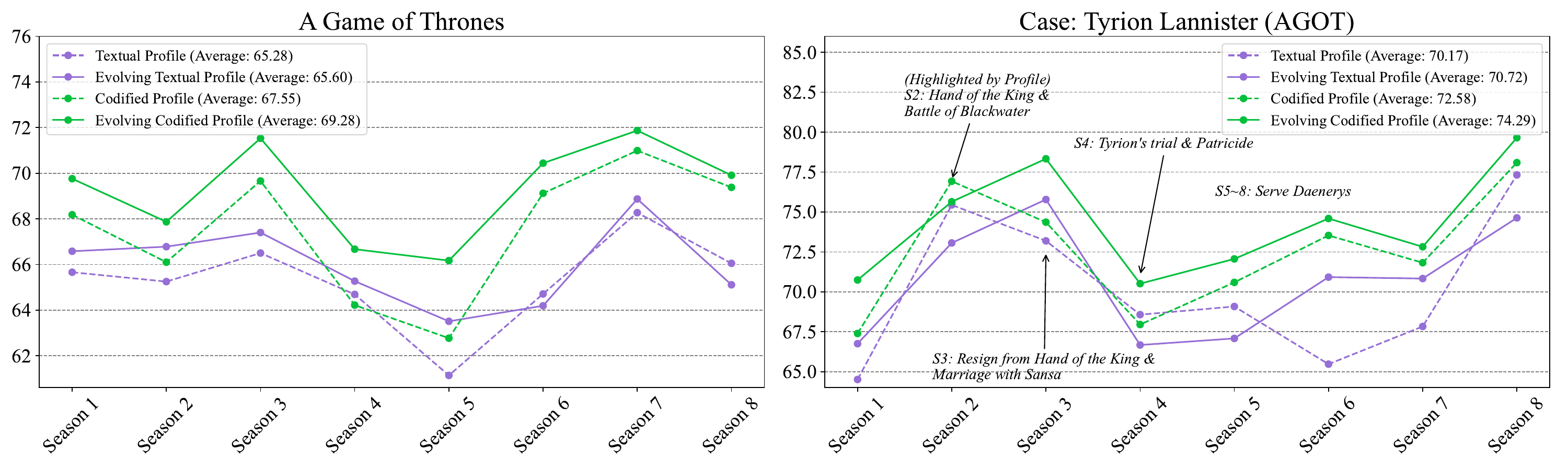}
  \vspace{-5mm}
    \caption{Role-playing performance with profiles evolving with the storyline. \textbf{Left:} Average role-playing performance on $8$ protagonists of \textit{``A Game of Thrones''}. \textbf{Right:} Case of role-playing as \textit{``Tyrion Lannister''} in \textit{``A Game of Thrones''} to analyze the working mechanism of profile evolving.}
  \vspace{-5mm}
    \label{fig:ecp_evolving}
\end{figure}

\paragraph{Evolving by Storyline} In Figure~\ref{fig:ecp_evolving}, we examine the effect of evolving codified profiles by the storyline. We select the $8$ protagonists (\textit{Tyrion}, \textit{Daenerys}, \textit{Cersei}, \textit{Jaime}, \textit{Arya}, \textit{Sansa}, \textit{Jon}, \textit{Bran}) in the full $8$ seasons of \textit{``A Game of Thrones''} because of AGOT's longest storyline among artifacts in the benchmark. There are $245.6$ scenes on average for each protagonist, solidifying the conclusion with massive data. The left subfigure verifies the benefit of evolving character logic along with the storyline. Codified profiles benefit more from evolving, which supports their advantage of directly revising the logic flow of characters.

The right subfigure presents a case study on character \textit{``Tyrion Lannister''} to specify how the evolving mechanism synchronizes characters with the storyline. Tyrion's profile generally centers around his highlighted moment when Tyrion successfully defends the capital in the Battle of Blackwater against the outnumbered troops of Stannis. This is also reflected in the subfigure that Tyrion is best grounded without evolving in Season 2 when the Battle of Blackwater occurs. However, in Season 3 and 4, just after the Battle of Blackwater, we observe a high demand for profile synchronization, even more urgent than in Season 5-8 when Tyrion escapes from the capital to serve Daenerys. This pattern suggests that the most critical need for evolving profiles arises in the aftermath, where characters undergo rapid development and shifting motivations. Codified profiles with evolution enable timely updates that reflect these transitions, ensuring sustained alignment between the character’s internal logic and narrative progression.

\begin{figure}[ht]
    \centering
    \includegraphics[width=1.0\linewidth]{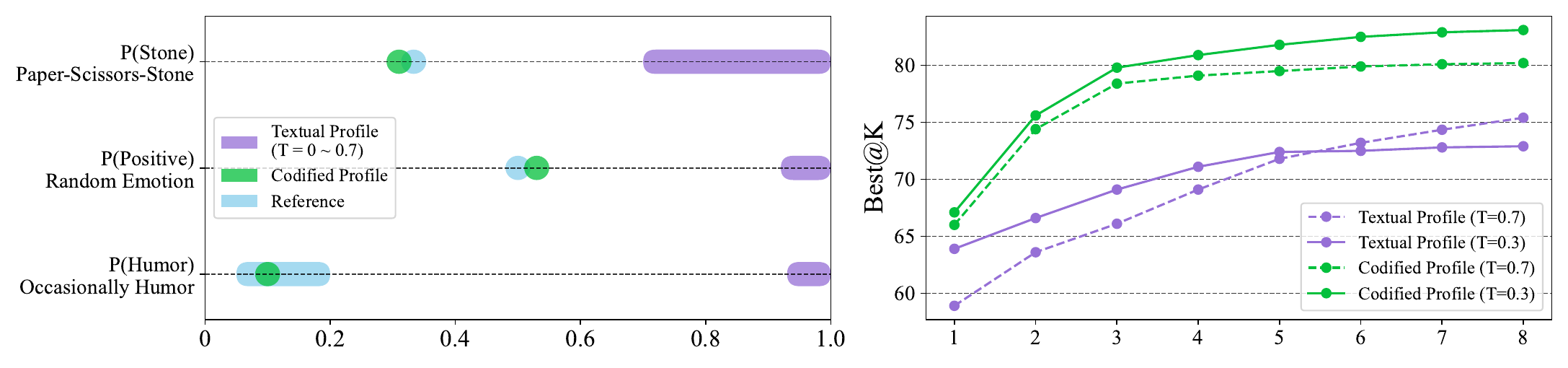}
  \vspace{-3mm}
    \caption{\textbf{Left:} Case experiments on controllable randomness simulation by textual and codified profiles. \textbf{Right:} Scenario coverage comparison, evaluated by Best@K performance under stochastic response setting.}
  \vspace{-3mm}
    \label{fig:ecp_randomness}
\end{figure}

\paragraph{Controllable Randomness} Figure~\ref{fig:ecp_randomness} plots the randomness control effect of textual and codified profiles in role-playing. The left subfigure includes $3$ simple randomness simulation cases in role-playing: (1) role-playing as a paper-scissor-stone robot that selects an action with equal probability in each turn; (2) role-playing as a robot that responds in positive or negative emotion with equal probability; (3) role-playing as a robot that occasionally includes humor with low probability in response. Prompting with textual profiles (with Temperature $0\sim 0.7$) shows significant bias in probability simulation as $P(\text{Stone})$, $P(\text{Positive})$, $P(\text{Humor})$ are all in high probability. In contrast, codifying probability can precisely control the randomness for the character to perform different responses with explicit probabilistic control flow. We show the codified profiles of the $3$ cases in Appendix~\ref{apdx:extended_case}, whose logic matches with the expected character logic.

The right subfigure shows the application of randomness simulation to complex characters in the Fandom benchmark. As expected, codified profiles implemented via explicit randomness in code control flow achieve stronger Best\@K performance, meaning they match ground-truth actions within K runs more often. Textual profiles rely on raising temperature to explore variability, which degrades single-run precision and introduces inconsistency. By decoupling behavioral exploration from sampling noise, codified profiles deliver both diversity and accuracy at low temperatures, making them diverse yet robust for high-quality role-playing.

\subsection{Smaller Role-playing LLMs}

\begin{wrapfigure}{r}{0.5\textwidth}
  \begin{center}
  \vspace{-5mm}
    \includegraphics[width=0.5\textwidth]{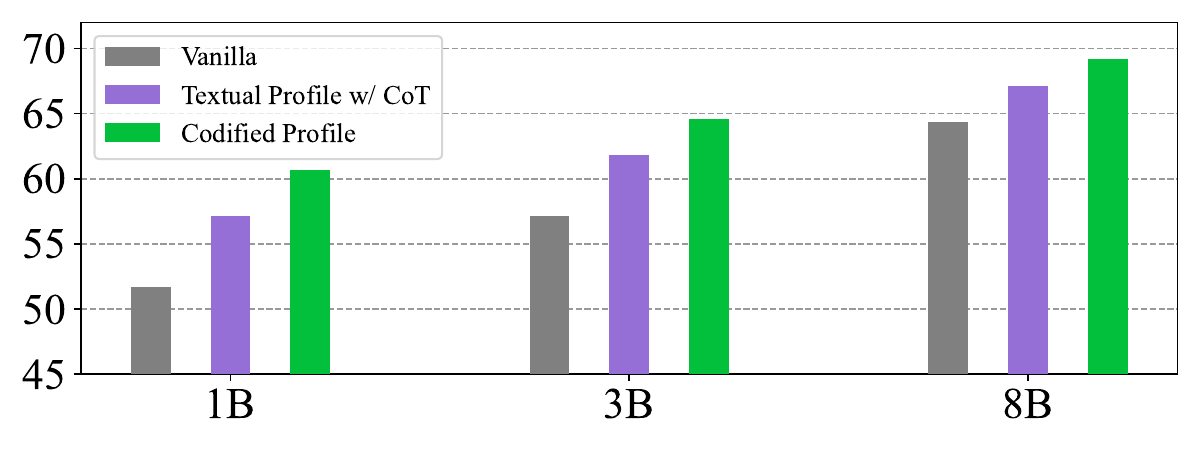}
  \end{center}
  \vspace{-3mm}
  \caption{The role-playing performance with smaller role-playing LLMs.}
  \label{fig:ecp_slm}
\end{wrapfigure}

We compare vanilla prompting, textual profiles with chain-of-thought, and codified profiles across LLaMA-3 models of 1B (3.2), 3B (3.2), and 8B (3.1) parameters in Figure~\ref{fig:ecp_slm}, and observe that the advantage of codified profiles grows as model size decreases. Notably, smaller LLMs with codified profiles rival the larger LLMs with textual profiles for both 1B and 3B cases. Manual inspection shows smaller models often fail to reconstruct character logic from free-form text but still perform well when logic is preprocessed by stronger LLMs (e.g. \texttt{gpt-4.1}) into explicit conditions. This supports our central claim in the introduction: codified profiles reduce reliance on the LLM’s implicit and inconsistent reasoning by enforcing structured, interpretable behavior, which is especially critical for smaller models with limited reasoning ability.

\begin{table}
\centering
\small
\scalebox{.99}{
\begin{tabular}{llcccccccc}
\toprule
\multicolumn{2}{l}{{Artifact}} & {Haruhi} & {K-On!} & {JOJO} & {FMA} & {AGOT} & {ATLA} & Average \\
\midrule
\multirow{4}*{\rotatebox{90}{Main}}
& 1B + Text & $51.60$ & $59.73$ & $52.76$ & $59.11$ & $54.25$ & $58.54$ & $56.00$ \\
& 1B + Code & $55.83$ & $60.80$ & $54.28$ & $60.32$ & $58.96$ & $60.37$ & $58.43$ \\
& 1B + Code + Distill & $62.25$ & $61.40$ & $55.01$ & $59.96$ & $60.09$ & $61.97$ & $\textbf{60.21}$ \\
\cmidrule(lr){2-9}
& 8B + Text & $70.53$ & $68.47$ & $60.64$ & $68.02$ & $61.52$ & $66.70$ & $65.98$ \\
\midrule
\multirow{4}*{\rotatebox{90}{Minor}} 
& 1B + Text & & $63.17$ & $61.98$ & $52.46$ & $58.04$ & $61.36$ & $59.40$ \\
& 1B + Code & & $64.98$ & $60.21$ & $60.68$ & $60.13$ & $66.97$ & $62.59$ \\
& 1B + Code + Distill & & $66.29$ & $61.15$ & $61.29$ & $62.74$ & $69.56$ & $\textbf{64.25}$ \\
\cmidrule(lr){2-9}
& 8B + Text & & $65.88$ & $64.70$ & $65.83$ & $66.21$ & $65.88$ & $65.70$ \\
\bottomrule
\end{tabular}
}
\vspace{2mm}
\caption{Codified profile and distilled 100M condition checker support high-quality role-playing based on 1B role-playing LLM.}
\label{tab:distill}
\end{table}

\paragraph{Distilled Condition Checker} As \texttt{check\_condition(scene, question)} is a discriminative task, we explore the use of a distilled lightweight classifier to improve both performance and efficiency. Instead of querying the role-playing model for every condition, we distill from \texttt{gpt-4.1}'s condition-checking outputs using 415 scenes (5 per character, $8\%$ of all scenes) and obtain 20,759 labeled discrimination cases. A 3-class \texttt{deberta-v3-base} model (0.1B)~\citep{debertav3} is trained on 90\% of the data for 5 epochs and achieves 70.53\% consistency with \texttt{gpt-4.1} on the held-out 10\%.

As shown in Table~\ref{tab:distill}, integrating this distilled discriminator into a 1B model with codified profiles leads to stronger role-playing performance than relying solely on the 1B LLM for both generation and condition checking. The 0.1B discriminator is over 10 times more efficient than the 1B LLM, reducing the overall execution cost to near negligible levels. This design enables a highly efficient 1B + Code + Distill system that approaches the performance of 8B + Text (even outperforming on minor characters in K-On! and ATLA), demonstrating the scalability of codified reasoning in resource-constrained settings.

This result highlights the practicality of further modularizing reasoning components within the codified profile framework. By offloading semantic condition evaluation to a small, specialized model, we significantly enhance the cost-effectiveness and accessibility of high-quality role-playing with smaller LLMs. The trade-off is the need to incorporate multi-model collaboration within the role-playing system, rather than relying solely on a single role-playing LLM. However, this modular design brings clear gains in both efficiency and performance, especially in low-resource settings like local deployment.

\section{Analysis}

\subsection{Segmentation Strategy}

\begin{wrapfigure}{r}{0.35\textwidth}
  \begin{center}
  \vspace{-5mm}
    \includegraphics[width=0.35\textwidth]{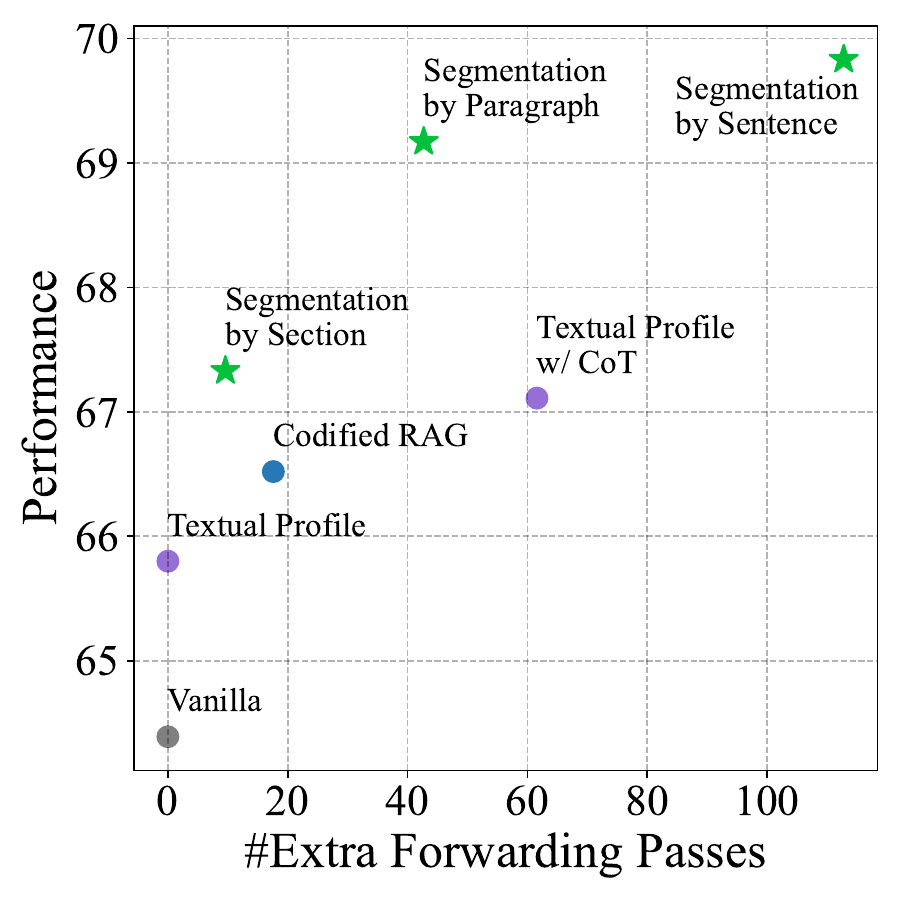}
  \end{center}
  \vspace{-5mm}
  \caption{The role-playing performance with different segmentation strategies.}
  \vspace{-10mm}
  \label{fig:ecp_segmentation}
\end{wrapfigure}

Figure~\ref{fig:ecp_segmentation} shows the effect of different segmentation strategies, comparing section-level, paragraph-level, and sentence-level codification. As the segmentation becomes more fine-grained, role-playing performance improves, with a clear gain when moving from section to paragraph. However, the improvement from paragraph to sentence is marginal and comes at a significant computational cost, nearly tripling the number of forward passes. This supports our design choice, as paragraphs naturally group logically related statements (at least for profiles in Fandom), making them an effective unit for codification. In contrast, section-level segmentation leads to information loss, while sentence-level segmentation introduces inference cost. Therefore, we adopt paragraph-level segmentation for the main experiments to balance fidelity and efficiency.

\subsection{Codification Performance} 

\begin{table}
\centering
\small
\scalebox{.95}{
\begin{tabular}{lccccccccc}
\toprule
\multirow{2}*{Grouped by} & \multicolumn{4}{c}{If-depth} & \multirow{2}*{w/ Branch} & \multirow{2}*{w/ Random} & \multirow{2}*{Personality} & \multirow{2}*{Ability} & \multirow{2}*{Relation} \\
\cmidrule(lr){2-5}
& $1$ & $2$ & $3$ & $4$ & \\
\midrule
Precision & $98.5$ & $96.0$ & $92.5$ & $93.8$ & $98.5$ & $91.0$ & $97.50$ & $92.50$ & $96.00$ \\
Recall & $94.0$ & $93.0$ & $89.5$ & $81.3$ & $95.0$ & $96.5$ & $96.00$ & $94.00$ & $98.50$ \\
Both & $94.0$ & $92.0$ & $87.0$ & $75.0$ & $95.0$ & $89.5$ & $95.50$ & $91.00$ & $95.00$ \\
\bottomrule
\end{tabular}
}
\vspace{2mm}
\caption{Manual evaluation on codification of profile segments grouped by different attributes.}
\label{tab:code_performance}
\end{table}

We manually analyze the codification ability in the preprocessing stage on $200$ cases grouped by different attributes (except for codes with if-depth $4$, which only have $16$ cases). Table~\ref{tab:code_performance} presents the manual evaluation of the codification performance of \texttt{gpt-4.1}, showing strong overall accuracy in translating natural language character descriptions into structured logic. Precision remains high across all settings, exceeding 98\% for shallow control structures and most attribute types, indicating that generated logic closely matches the intended behavior. While recall slightly decreases as the if-depth increases, dropping to 81.3\% at depth 4, this reflects the growing complexity of deeper nesting rather than fundamental failure. Notably, codification is robust across personality, ability, and relational traits, achieving over 95\% F1 in most categories. These results demonstrate that \texttt{gpt-4.1} is highly capable of converting rich character profiles into executable logic with minimal loss of fidelity.

\subsection{Case Study}

\begin{figure}[ht]
    \centering
    \includegraphics[width=1.0\linewidth]{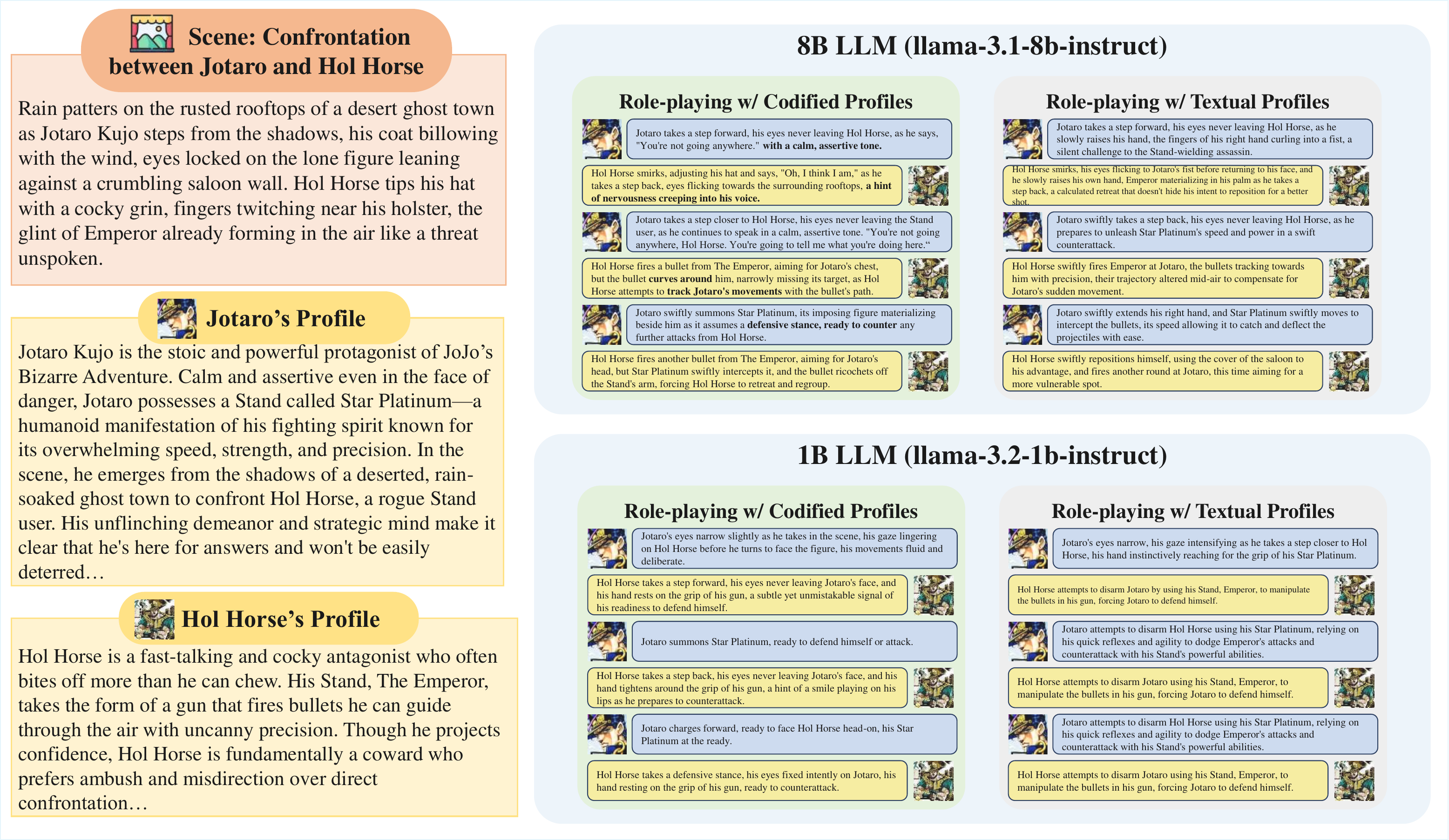}
  \vspace{-3mm}
    \caption{Case study of multi-turn role-playing in a given scene: \textit{``Jotaro confronting Hol Horse.'' (JOJO's Bizarre Adventure)}}
  \vspace{-3mm}
    \label{fig:case_study}
\end{figure}

Figure~\ref{fig:case_study} illustrates a case study that highlights the advantage of codified profiles in a direct role-play scenario. We construct a dynamic scene featuring a standoff between Jotaro and Hol Horse from \textit{``JoJo's Bizarre Adventure''}, and prompt LLMs to role-play as each character. Both textual and codified profiles are evaluated using \texttt{llama-3.1-8b-instruct} and \texttt{llama-3.2-1b-instruct} to compare performance across model scales and profile formats. For both characters,

The codified profile produces more coherent and character-faithful role-playing by enforcing consistent personality traits and logically grounded actions. For example, when Hol Horse fires a bullet and misses, the codified logic immediately triggers Jotaro to summon Star Platinum in defense, an explicit and reactive sequence that mirrors the canon. In contrast, the textual profile drifts into generic exchanges, with Jotaro raising a fist and stepping back without clear narrative consequence. By using explicit condition checks and control flow, the codified approach reduces reliance on implicit model reasoning and delivers more believable, interactive, and narratively aligned responses.

The codified profile enables smaller LLMs to produce more coherent and context-aware role-playing by offloading complex reasoning into structured logic. Unlike the textual profile, which degenerates into repetitive outputs such as "Hol Horse attempts to disarm Jotaro" without meaningful variation, the codified version guides the 1B model through a reactive and evolving exchange. For example, when Hol Horse signals aggression, Jotaro immediately summons Star Platinum, leading to a logical sequence of retreat and confrontation. This explicit control flow allows the smaller model to maintain character fidelity and narrative coherence that would otherwise require far larger model capacity.

We further include extended case studies for codified profile workflow in Appendix~\ref{apdx:extended_case}, which can be referred to understand the mechanism and advantage of codified profiles.

\section{Conclusion}

We propose Codified Profiles, which offer a principled and scalable framework for character role-playing by compiling behavior into interpretable, executable functions. Our experiments demonstrate that this approach outperforms prompt-based methods in consistency, adaptability, and controllable variability, while enabling strong performance even in small models. These findings highlight Codified Profiles as a practical foundation for building efficient, locally deployable role-play agents across a wide range of applications. Future work, as discussed in Appendix~\ref{apdx:limitation}, includes extending codification with richer operators and helper tools, modeling environmental logic for world-level role-play, and exploring the scope of codifiable textual content.

\section*{Acknowledgement}

This work aims to contribute not only to the research community but also to a broader ACG community by introducing more powerful role-playing agents. It is also done in memory of the 17th \emph{Koishi's Day} (May 14th), 2025, since the release of TH11, Touhou Chireiden $\sim$ Subterranean Animism\footnote{\href{https://en.wikipedia.org/wiki/Subterranean_Animism}{https://en.wikipedia.org/wiki/Subterranean\_Animism}} in 2008.

\bibliographystyle{neurips_2025}
\bibliography{neurips_2025}



\clearpage

\appendix

\section{Limitation and Future Work}
\label{apdx:limitation}

As a pioneering study that integrates executable code into character-based role-playing, this paper focuses on presenting the broad applicability and core benefits of codified profiles. However, several promising further explorations are remained for future works.

\paragraph{More operators and helper functions} First, while our framework centers on the \texttt{check\_condition} helper function due to its semantic clarity and general applicability, future work can introduce additional operators and helper tools to support more complex behaviors. For example, in role-playing game (RPG) scenarios, codified profiles could include arithmetic operations for managing health or damage, procedures for activating skills, or functions for spatial movement and positioning. These extensions would enable more dynamic and interactive role-play beyond conditional responses.

\paragraph{World-level role-playing} Second, codification can be expanded beyond individual characters to represent the fictional world itself. Codifying environmental rules, societal structures, or magical systems would allow models to simulate consistent world behavior and responses. While this paper does not pursue world-level codification due to the lack of benchmarks for evaluating environmental reactivity, it presents a promising direction for future research in narrative simulation.

\paragraph{Scope of codification} Finally, codification introduces a potential risk of information loss by abstracting nuanced textual descriptions into structured logic. We view this as a trade-off that can be managed: codified functions can always reference the original profile when needed, and conservative codification that only translates explicit control flows already improves performance. Developing hybrid strategies that balance interpretability and fidelity remains an important direction for making codified profiles more robust and expressive.

These limitations reflect deliberate design choices to focus on a broad and extensible framework. We hope they inspire future work that builds on codified reasoning to support richer, more controllable, and scalable role-playing systems.

\section{Prompts and Templates}
\label{apdx:prompt_template}

\begin{figure}
    \centering
    \includegraphics[width=0.9\linewidth]{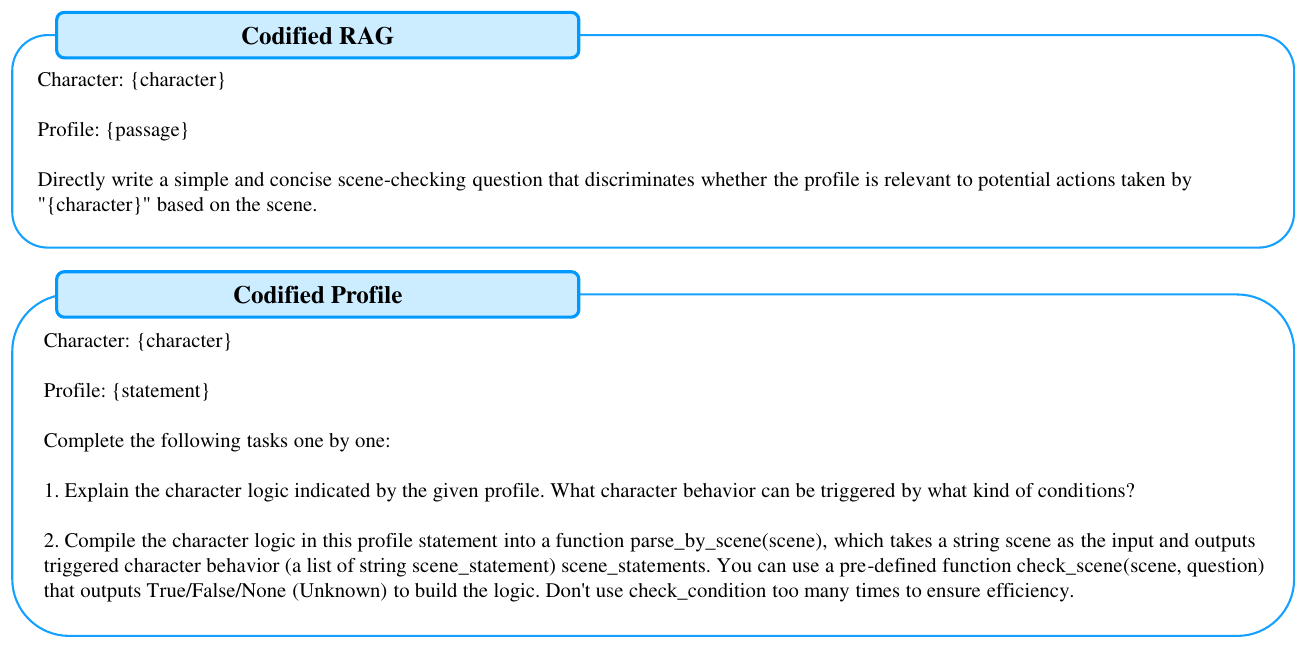}
    \caption{The preprocessing prompts used in our experiments.}
    \label{fig:prompt}
\end{figure}

\begin{figure}
    \centering
    \includegraphics[width=0.9\linewidth]{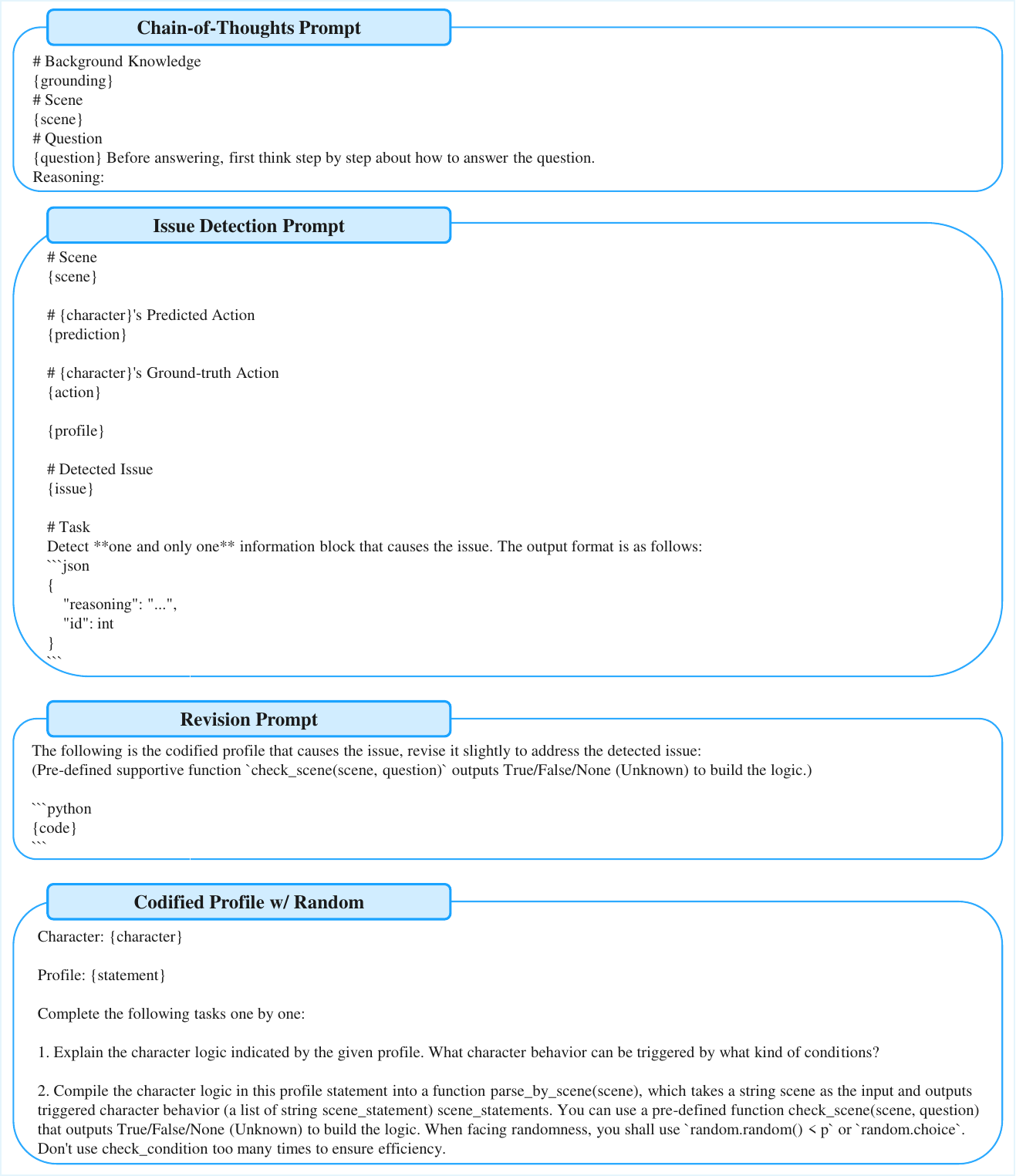}
    \caption{The shared prompts used for evolving profile and scholastic response.}
    \label{fig:shared_evolving}
\end{figure}

\begin{figure}
    \centering
    \includegraphics[width=0.9\linewidth]{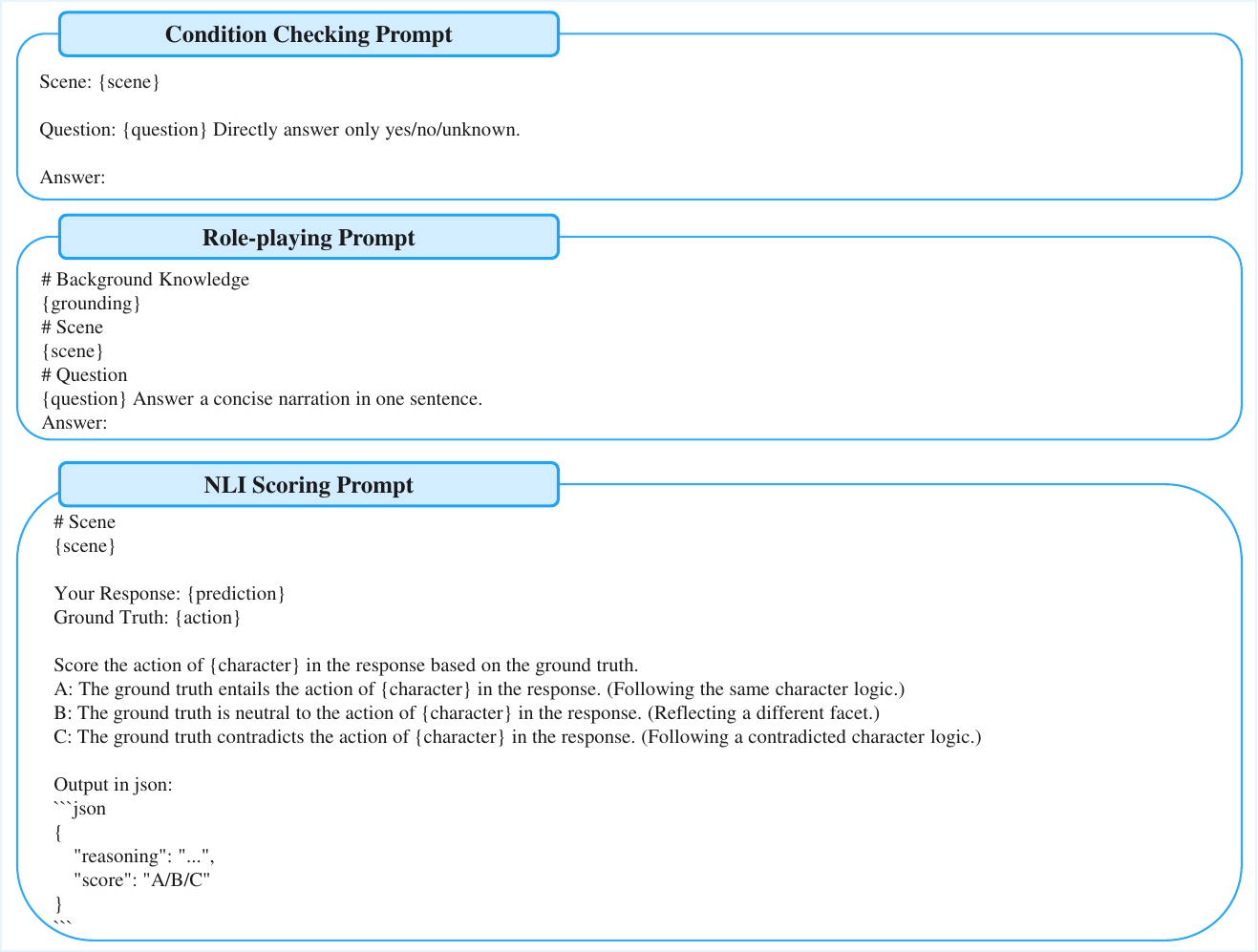}
    \caption{The shared prompts used in our experiments.}
    \label{fig:shared_prompt}
\end{figure}

Figures from~\ref{fig:prompt} to~\ref{fig:shared_prompt} show the prompts used in our experiments for result reproduction.

\clearpage

\section{NLI Metric Reliability}
\label{apdx:nli_reliability}

We manually check $5$ NLI scoring cases for all $83$ characters, and find $92.35\%$ entailed, $90.91\%$ neutral, and $88.89\%$ contradicted judgments align with human evaluation, which validates the usage of LLM for automated NLI scoring.

\section{Character-wise Comparison}
\label{apdx:character_comparison}

\begin{table}
\centering
\small
\scalebox{.99}{
\begin{tabular}{cccccccc}
\toprule
\multicolumn{4}{c}{Best Performance} & \multicolumn{4}{c}{Worst Performance} \\
\cmidrule(lr){1-4}
\cmidrule(lr){5-8}
Rank & Character & Win Rate & Artifact & Rank & Character & Win Rate & Artifact \\
\midrule
1 & Sansa & 81.25 & AGOT & 60 & Winry & 32.00 & FMA \\
2 & Tywin & 70.00 & AGOT & 59 & Mugi & 34.15 & K-On! \\
3 & Littlefinger & 68.42 & AGOT & 58 & Riza & 34.78 & FMA \\
4 & Joffrey & 67.86 & AGOT & 57 & Stannis & 35.00 & AGOT \\
5 & Jon & 67.57 & AGOT & 56 & Iroh & 38.18 & ATLA \\
6 & D'Arby & 64.71 & JOJO & 55 & Bran & 38.89 & AGOT \\
7 & Nagato & 63.64 & Haruhi & 54 & Koizumi & 41.18 & Haruhi \\
8 & Hol Horse & 62.07 & JOJO & 53 & Sandor & 42.86 & AGOT \\
9 & Lust & 61.54 & FMA & 52 & Joseph & 42.86 & JOJO \\
10 & Iggy & 60.71 & JOJO & 51 & DIO & 44.68 & JOJO \\
\bottomrule
\end{tabular}
}
\vspace{2mm}
\caption{Win rates of codified profile against textual profile.}
\label{tab:character_performance}
\end{table}

Table~\ref{tab:character_performance} presents a character-level (characters with at least $20$ scenes) comparison of win rates between codified and textual profiles (ties are removed), highlighting which characters benefited most from codification. Notably, the top-performing characters such as Tywin, Littlefinger, and D'Arby all exhibit subtle, strategic, or context-sensitive behavior in the source material. These ``fickle characters'' often require conditional responses based on shifting alliances, hidden motives, or indirect manipulation, making them well-suited to codified logic that explicitly encodes behavioral rules and scenario-based decision-making. In contrast, characters with lower win rates, such as Winry, Mugi, and Riza, tend to exhibit more emotionally driven or straightforward behavior, where textual prompting may suffice and codification provides less added benefit. This pattern suggests that codified profiles are especially advantageous for characters whose actions depend on nuanced reasoning and consistent application of complex behavioral constraints. Such discovery further shows a complementing relation between textual and codified profiles, explaining why the ensemble between them generally works.

\section{Extended Case Study}
\label{apdx:extended_case}

\begin{figure}
    \centering
    \includegraphics[width=0.99\linewidth]{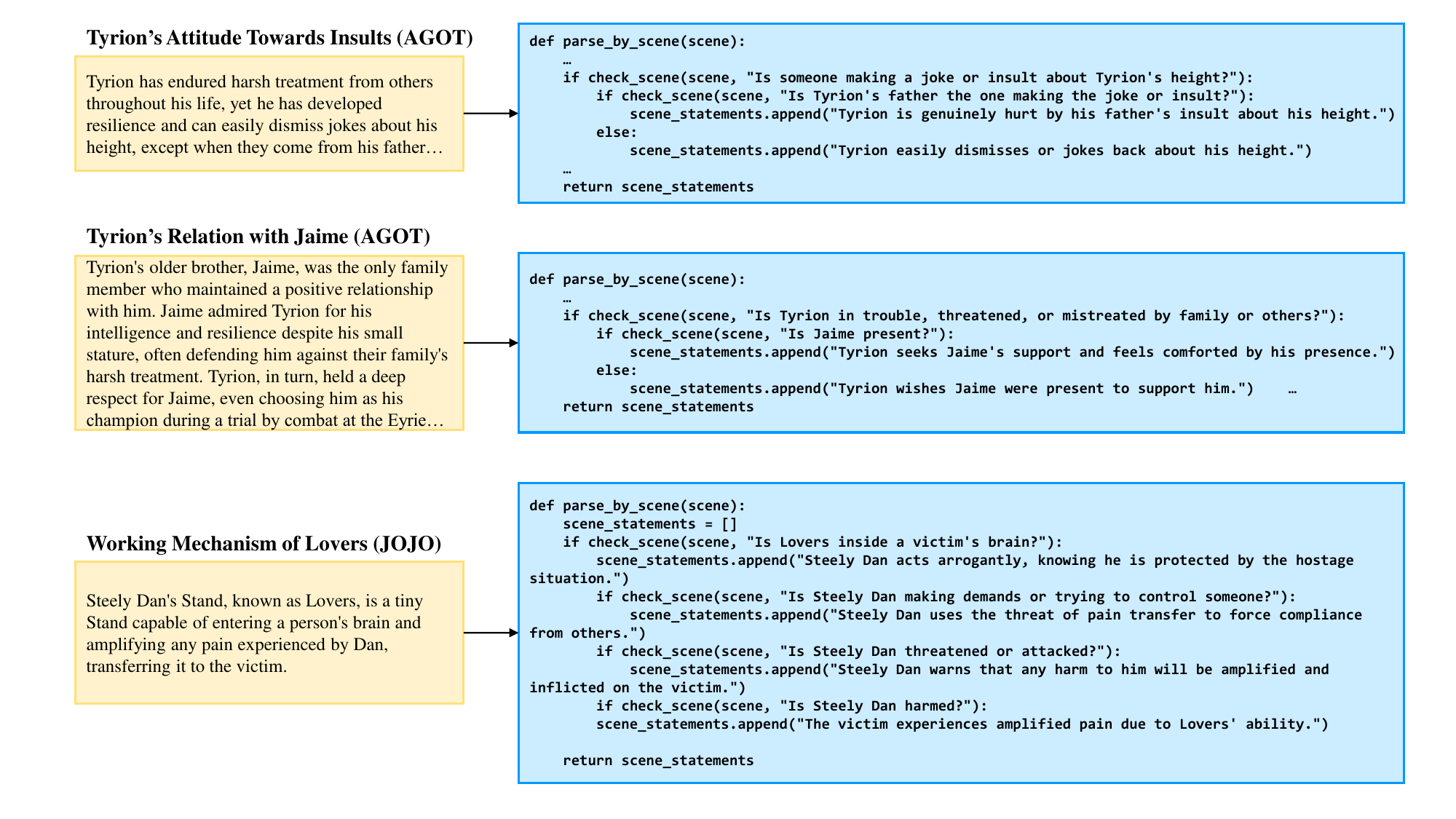}
    \caption{Codification cases for personality, relation, and working mechanism of superpower.}
    \label{fig:extended_case_codification}
\end{figure}

\begin{figure}
    \centering
    \includegraphics[width=0.99\linewidth]{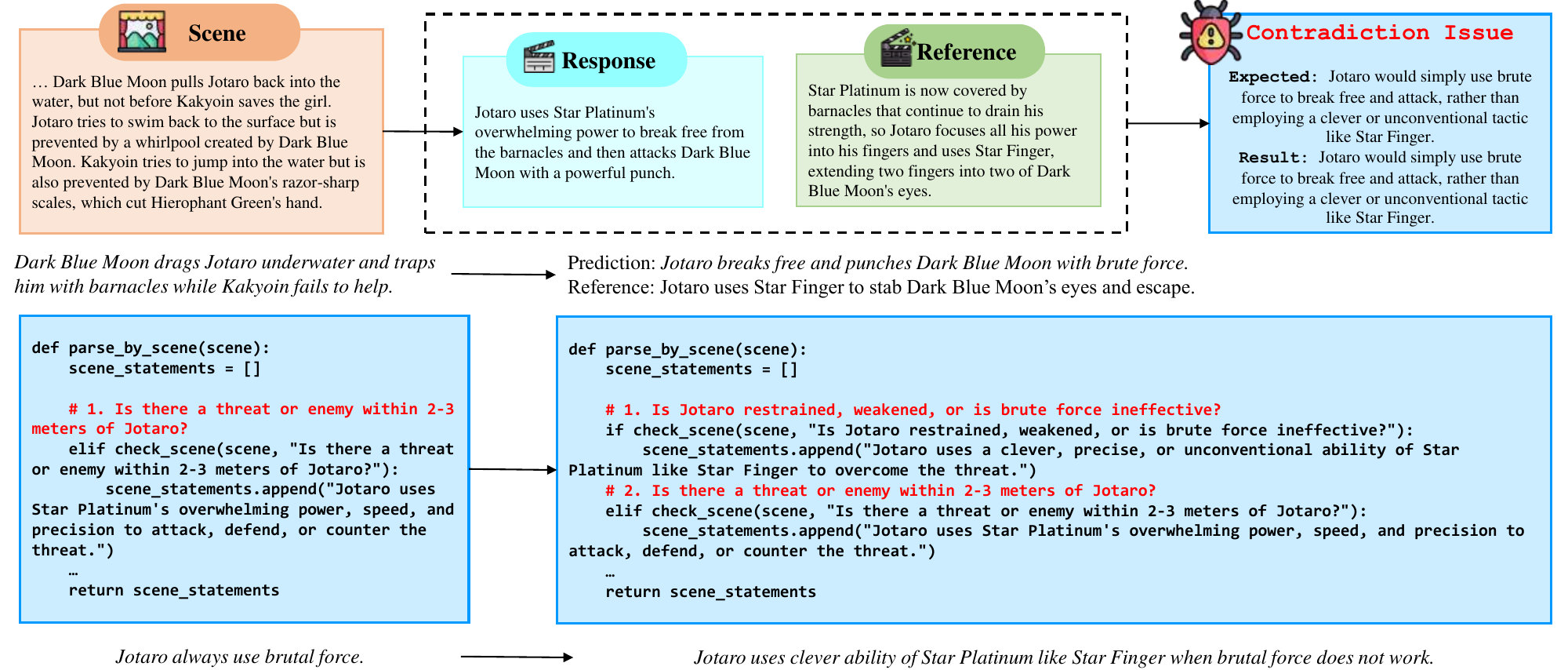}
    \caption{A case for the evolving mechanism for codified profiles.}
    \label{fig:extended_case_reflection}
\end{figure}

\begin{figure}
    \centering
    \includegraphics[width=0.99\linewidth]{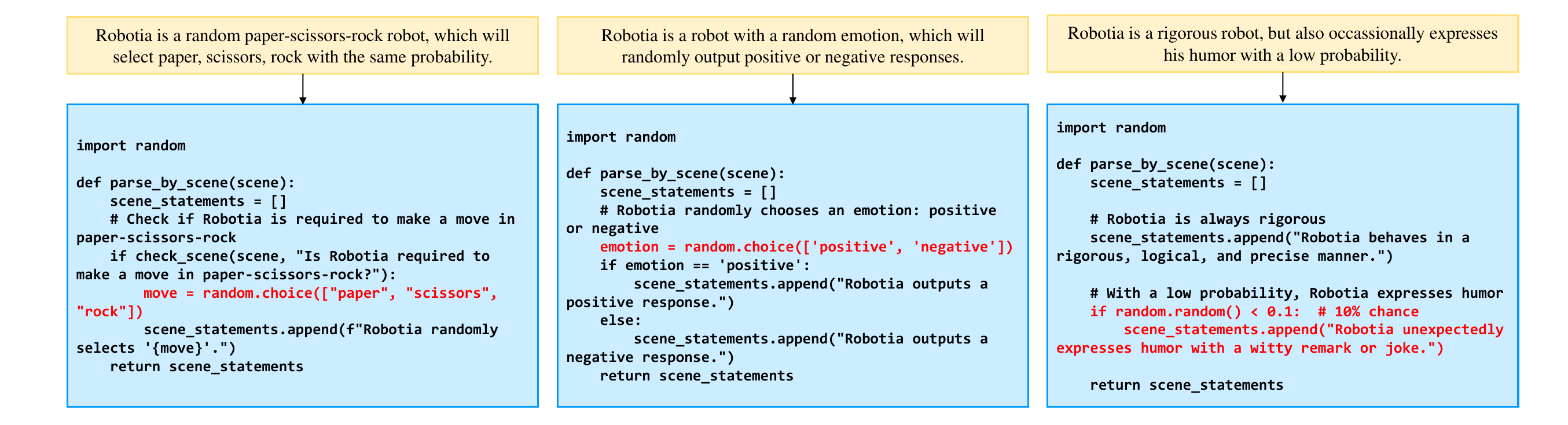}
    \caption{Codification cases with randomness simulation involved.}
    \label{fig:extended_case_random}
\end{figure}

\paragraph{Codification} In Figure~\ref{fig:extended_case_codification}, we showcase several codification workflow, including personality, relation and working mechanism of superpower. The cases validate the ability of codified profile to reflect exceptions (\textit{``except when insults come from his father.)''}), situational thoughts (\textit{``Is Jaime present?''}), and superpower checking (\textit{``Dan is harmed''}$\rightarrow$\textit{``The victim is harmed.''}|\textit{``Lovers inside a victim's brain''}). These cases provide a straightforward illustration on how codified profile builds up the character logic.

\paragraph{Evolving} An evolving case about Jotaro's fighting strategy is presented in Figure~\ref{fig:extended_case_reflection}, which shows a mismatch between the predicted and reference actions when he is trapped underwater by Dark Blue Moon. The initial codified logic assumed Jotaro always resorts to brute force, resulting in an incorrect prediction where he punches to escape. However, the reference reveals that Jotaro uses the more precise Star Finger technique to blind the enemy and break free. Based on this contradiction, the profile is revised to reflect a more accurate strategy: Jotaro prefers clever abilities like Star Finger when brute strength is ineffective. This update enables the codified profile to better align with his adaptive combat behavior in future scenes.

\paragraph{Controllable Randomness} We present the specific codes to demonstrate how codified profiles enable controllable randomness in character behavior. In the first, Robotia plays paper-scissors-rock, with each move selected uniformly using \texttt{random.choice}, ensuring balanced randomness. In the second, Robotia expresses either a positive or negative emotion with equal probability, highlighting how codified logic maintains behavioral diversity more reliably than prompt-based sampling. In the third, Robotia consistently behaves logically but has a 10\% chance of adding humor, showing how low-probability actions can be precisely embedded without disrupting core traits. These examples demonstrate how codified randomness offers fine-grained, interpretable variability beyond what temperature tuning alone can provide.

\section{Character Information}
\label{apdx:character_info}

\begin{figure}
    \centering
    \includegraphics[width=0.99\linewidth]{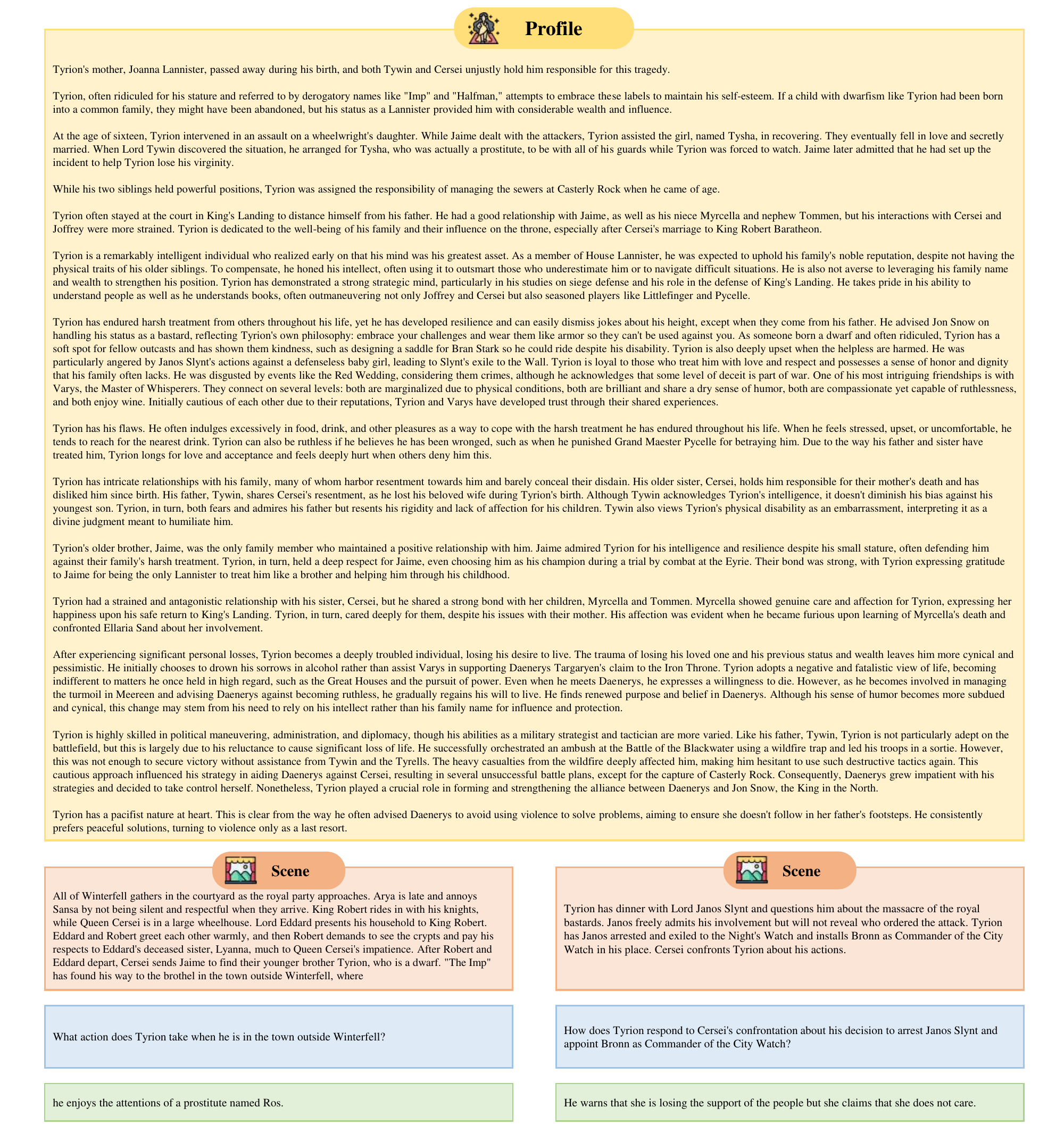}
    \caption{An example of a profile and testing cases used in our experiments. (Tyrion Lannister)}
    \label{fig:data_point}
\end{figure}

In Tables~\ref{tab:character_info} and~\ref{tab:character_info_2}, we present brief introductions of characters involved in our experiments for readers who are unfamiliar with these artifacts. In Figure~\ref{fig:data_point}, we also showcase a profile and testing cases for better clarity of our benchmark.

\begin{table}
\centering
\small
\scalebox{.7}{
\begin{tabular}{cccp{15cm}}
\toprule
\multirow{10}*{\rotatebox{90}{Haruhi}} & \multirow{10}*{\rotatebox{90}{main}} & Haruhi & Haruhi is an eccentric and energetic high school student whose curiosity and unconventional outlook drive the extraordinary events of the "Haruhi" series. \\
 &  & Kyon & Kyon is a witty and pragmatic high school student who serves as the narrator and reluctant companion to the eccentric Haruhi Suzumiya in the "Haruhi Suzumiya" series. \\
 &  & Nagato & Nagato Yuki is a quiet, enigmatic member of the SOS Brigade in the "Haruhi Suzumiya" series, known for her extraordinary intelligence and mysterious origins. \\
 &  & Koizumi & Koizumi is a mysterious and ever-smiling transfer student in the "Haruhi Suzumiya" series, who serves as an esper and a key member of the SOS Brigade. \\
 &  & Asahina & Asahina Mikuru is a shy and gentle upperclassman in the "Haruhi Suzumiya" series, often roped into the SOS Brigade's antics as their adorable and mysterious "mascot." \\
 \midrule
\multirow{18}*{\rotatebox{90}{K-On!}} & \multirow{10}*{\rotatebox{90}{main}} & Yui & Yui Hirasawa is the cheerful and airheaded lead guitarist of the high school band in the anime "K-On!", known for her infectious enthusiasm and love of sweets. \\
 &  & Ritsu & Ritsu Tainaka is the energetic and playful drummer of the high school band in the anime "K-On!" known for her mischievous antics and close friendship with her bandmates. \\
 &  & Mio & Mio Akiyama is the shy and talented bassist of the high school band in the anime "K-On!", known for her gentle personality and musical prowess. \\
 &  & Mugi & Mugi, whose full name is Tsumugi Kotobuki, is a gentle and cheerful keyboardist in the anime "K-On!", known for her wealth, kindness, and love of sharing sweets with her friends. \\
 &  & Azusa & Azusa Nakano is a diligent and talented guitarist who joins the light music club in the anime "K-On!", quickly becoming an integral and endearing member of the group. \\
\cmidrule(lr){2-4}
 & \multirow{8}*{\rotatebox{90}{minor}} & Sawako & Sawako is a shy and soft-spoken high school girl who gradually opens up to her classmates in the heartwarming series "Kimi ni Todoke." \\
 &  & Nodoka & Nodoka is a gentle and bookish high school student from the anime "K-On!", known for her close friendship with Yui Hirasawa and her responsible role as a member of the student council. \\
 &  & Ui & Ui is a kind-hearted and responsible younger sister in "K-On!", known for her maturity and unwavering support for her older sister, Yui. \\
 &  & Jun & Jun is a thoughtful and reserved member of the group in 'Kon,' known for their quiet intelligence and unwavering loyalty to their friends. \\
 \midrule
\multirow{24}*{\rotatebox{90}{Fullmetal Alchemist}} & \multirow{10}*{\rotatebox{90}{main}} & Edward & Edward Elric is a brilliant and determined young alchemist who embarks on a perilous journey to restore his and his brother’s bodies after a failed alchemical experiment in "Fullmetal Alchemist." \\
 &  & Alphonse & Alphonse Elric is a gentle and kind-hearted young alchemist whose soul is bound to a towering suit of armor after a failed alchemical ritual in "Fullmetal Alchemist." \\
 &  & Winry & Winry Rockbell is a talented automail engineer and childhood friend of the Elric brothers in "Fullmetal Alchemist," known for her mechanical expertise and compassionate nature. \\
 &  & Roy & Roy Mustang is a charismatic and ambitious State Alchemist in "Fullmetal Alchemist," renowned for his mastery of flame alchemy and his unwavering determination to reform the military from within. \\
 &  & Ling & Ling Yao is a charismatic and ambitious prince from Xing in "Fullmetal Alchemist," driven by his quest for immortality and a deep sense of responsibility toward his people. \\
\cmidrule(lr){2-4}
 & \multirow{14}*{\rotatebox{90}{minor}} & Envy & Envy is a cunning and sadistic homunculus from Fullmetal Alchemist, known for their shapeshifting abilities and deep-seated resentment toward humanity. \\
 &  & Izumi & Izumi Curtis is a fiercely skilled alchemist and martial artist who serves as a tough but caring mentor to the Elric brothers in Fullmetal Alchemist. \\
 &  & Lust & Lust, one of the seven Homunculi in "Fullmetal Alchemist," is a cunning and seductive antagonist known for her deadly extendable fingers and her complex, enigmatic motivations. \\
 &  & Scar & Scar is a vengeful and enigmatic warrior in "Fullmetal Alchemist," driven by the trauma of his war-torn past and a mission to punish State Alchemists for their role in the destruction of his people. \\
 &  & Greed & Greed is a cunning and charismatic homunculus from Fullmetal Alchemist, driven by an insatiable desire for material possessions, power, and immortality, yet harboring a surprisingly complex sense of loyalty and independence. \\
 &  & Riza & Riza Hawkeye is a highly skilled sharpshooter and the loyal lieutenant to Colonel Roy Mustang in Fullmetal Alchemist, known for her calm demeanor, unwavering sense of duty, and deep sense of loyalty. \\
 &  & King Bradley & King Bradley, also known as Wrath, is the enigmatic and fearsome leader of Amestris in "Fullmetal Alchemist," concealing his true identity as a deadly Homunculus. \\
 \midrule
\multirow{32}*{\rotatebox{90}{JOJO's Bizarre Adventure}} & \multirow{14}*{\rotatebox{90}{main}} & Jotaro & Jotaro Kujo is a stoic and powerful high school student who serves as the protagonist of "JoJo's Bizarre Adventure: Stardust Crusaders," renowned for his iconic Stand, Star Platinum, and his unyielding resolve. \\
 &  & Polnareff & Jean Pierre Polnareff is a brave and flamboyant French swordsman who joins the Joestar group in "JoJo's Bizarre Adventure: Stardust Crusaders," wielding the Stand Silver Chariot in his quest for justice and revenge. \\
 &  & Joseph & Joseph Joestar is a quick-witted and flamboyant protagonist from "JoJo's Bizarre Adventure," known for his clever tactics, brash personality, and signature catchphrase, "Your next line is..." \\
 &  & DIO & DIO is a charismatic and ruthless vampire antagonist from the "JoJo's Bizarre Adventure" series, known for his overwhelming power, cunning intellect, and iconic catchphrase, "Za Warudo!" \\
 &  & Kakyoin & Noriaki Kakyoin is a cool and intelligent Stand user who joins Jotaro Kujo and his friends on their journey in "JoJo's Bizarre Adventure: Stardust Crusaders," wielding the emerald-shooting Stand, Hierophant Green. \\
 &  & Avdol & Mohammed Avdol is a wise and loyal Stand user from Egypt, known for his fiery Stand Magician's Red and his unwavering support of Jotaro and his friends in "JoJo's Bizarre Adventure: Stardust Crusaders." \\
 &  & Iggy & Iggy is a small, scrappy Boston Terrier with a bad attitude and a love for coffee-flavored gum, who joins the Joestar group as a Stand user in "JoJo's Bizarre Adventure: Stardust Crusaders." \\
 & \multirow{18}*{\rotatebox{90}{minor}} & Hol Horse & Hol Horse is a cunning and flamboyant gunslinger Stand user from "JoJo's Bizarre Adventure: Stardust Crusaders," known for wielding the sentient revolver Stand, Emperor. \\
 &  & Alessi & Alessi is a minor antagonist from "JoJo's Bizarre Adventure: Stardust Crusaders," known for his cowardly demeanor and his Stand, Sethan, which has the power to regress people into younger versions of themselves. \\
 &  & D'Arby & D'Arby is a cunning and manipulative gambler from "JoJo's Bizarre Adventure," known for his deadly games of chance and his ability to steal souls from those who lose to him. \\
 &  & Steely Dan & Steely Dan is a cunning and sadistic Stand user from "JoJo's Bizarre Adventure: Stardust Crusaders," known for his manipulative tactics and his Stand, Lovers, which allows him to infiltrate and control the minds of his enemies. \\
 &  & Vanilla Ice & Vanilla Ice is a ruthless and fanatically loyal servant of DIO in "JoJo's Bizarre Adventure: Stardust Crusaders," wielding the deadly Stand Cream, which can erase anything it touches from existence. \\
 &  & Enya & Enya the Hag is a cunning and malevolent Stand user in "JoJo's Bizarre Adventure: Stardust Crusaders," serving as a loyal follower of Dio and wielding the deadly Stand Justice. \\
 &  & Oingo & Oingo is a mischievous Stand user from "JoJo's Bizarre Adventure: Stardust Crusaders," known for his ability to shapeshift his appearance using his Stand, Khnum, and his comical partnership with his brother Boingo. \\
 &  & Pet Shop & Pet Shop is a menacing, highly intelligent falcon who serves as the guardian of DIO's mansion in "JoJo's Bizarre Adventure: Stardust Crusaders," wielding the deadly Stand Horus. \\
 &  & Boingo & Boingo is a timid and eccentric Stand user from "JoJo's Bizarre Adventure: Stardust Crusaders," known for his prophetic comic book Stand, Tohth, which predicts the future in bizarre and often literal ways. \\
\bottomrule
\end{tabular}
}
\vspace{2mm}
\caption{Information of characters in our experiments (1/2).}
\label{tab:character_info}
\end{table}

\begin{table}
\centering
\small
\scalebox{.7}{
\begin{tabular}{cccp{15cm}}
\toprule
\multirow{50}*{\rotatebox{90}{A Game of Thrones}} & \multirow{16}*{\rotatebox{90}{main}} & Tyrion & Tyrion Lannister, the sharp-witted and sharp-tongued youngest son of Lord Tywin, navigates the treacherous politics of Westeros with cunning and humor despite being scorned for his stature. \\
 &  & Daenerys & Daenerys Targaryen, the exiled princess of the fallen Targaryen dynasty, begins her journey in "A Game of Thrones" as a timid young girl sold into marriage, destined to become a powerful and determined leader. \\
 & & Cersei & Cersei Lannister is the ambitious and cunning queen of the Seven Kingdoms, known for her beauty, ruthlessness, and fierce devotion to her family. \\
 &  & Jaime & Jaime Lannister, known as the Kingslayer, is a skilled and charismatic knight of the Kingsguard, infamous for killing the Mad King and renowned for his striking looks and complicated loyalties. \\
 &  & Robb & Robb Stark is the eldest son of Eddard and Catelyn Stark, a dutiful and honorable young lord who shoulders great responsibility as heir to Winterfell. \\
 &  & Eddard & Eddard Stark, the honorable and steadfast Lord of Winterfell, serves as Warden of the North. \\
 &  & Arya & Arya Stark is the fiercely independent and adventurous youngest daughter of Eddard and Catelyn Stark. \\
 &  & Catelyn & Catelyn Stark is the resolute and fiercely protective lady of Winterfell, whose loyalty to her family shapes her every action. \\
 &  & Sansa & Sansa Stark is the eldest daughter of Eddard and Catelyn Stark, known for her beauty, courtesy, and romantic dreams. \\
 &  & Jon & Jon Snow is the brooding, illegitimate son of Eddard Stark, raised at Winterfell and haunted by questions of identity and belonging. \\
 &  & Bran & Bran Stark is the curious and adventurous seven-year-old son of Eddard Stark, whose life changes forever after a fateful fall. \\
\cmidrule(lr){2-4}
 & \multirow{30}*{\rotatebox{90}{minor}} & Tywin & Tywin Lannister is the formidable and calculating head of House Lannister, renowned for his ruthless political acumen and unyielding pursuit of power. \\
 &  & Varys & Varys, known as the Spider, is the enigmatic and cunning Master of Whisperers in King’s Landing, weaving a vast web of spies throughout the Seven Kingdoms. \\
 &  & Joffrey & Joffrey Baratheon is the arrogant and cruel crown prince of the Seven Kingdoms known for his golden hair, petulant demeanor, and penchant for sadistic behavior. \\
 &  & Theon & Theon Greyjoy is the charismatic and cocky heir of the Iron Islands, serving as a ward of House Stark in Winterfell. \\
 &  & Stannis & Stannis Baratheon, the stern and unyielding younger brother of King Robert, is a relentless claimant to the Iron Throne. \\
 &  & Littlefinger & Petyr Baelish, known as Littlefinger, is a cunning and ambitious master of coin at King’s Landing, renowned for his manipulative schemes and silver tongue. \\
 &  & Melisandre & Melisandre, a mysterious and powerful priestess from Asshai, serves Stannis Baratheon wielding shadowy magic in the name of her fiery god, R'hllor. \\
 &  & Jorah & Jorah Mormont is a disgraced knight from Bear Island who serves as an exiled advisor to Daenerys Targaryen. \\
 &  & Sandor & Sandor Clegane, known as "the Hound," is a fearsome and brutally honest knight with a burned, disfigured face who serves as the bodyguard to Prince Joffrey Baratheon. \\
 &  & Shae & Shae is a young and alluring camp follower who becomes Tyrion Lannister’s lover during his time in the Lannister army. \\
 &  & Margaery & Margaery Tyrell is the clever and beautiful daughter of House Tyrell, renowned for her political acumen and poised ambition in the courtly intrigues of Westeros. \\
 &  & Davos & Davos Seaworth, known as the "Onion Knight," is a former smuggler turned loyal and honest advisor to Stannis Baratheon. \\
 &  & Renly & Renly Baratheon is the charismatic and handsome youngest brother of King Robert Baratheon, known for his charm, wit, and political ambition. \\
 &  & Bronn & Bronn is a cunning and pragmatic sellsword known for his sharp wit, deadly skill with a sword, and willingness to fight for the highest bidder. \\
 &  & Brienne & Brienne of Tarth, a formidable and loyal warrior known for her unwavering honor and unconventional appearance, is introduced as a noblewoman who defies traditional gender roles in pursuit of knighthood and justice. \\
 &  & Barristan & Barristan Selmy, known as Barristan the Bold, is a legendary and honorable knight of the Kingsguard serving King Robert Baratheon. \\
 &  & Mance & Mance Rayder, once a sworn brother of the Night’s Watch, is the charismatic and cunning King-Beyond-the-Wall who unites the wildlings. \\
 &  & Craster & Craster is a cruel and incestuous wildling who lives north of the Wall, notorious for marrying his own daughters and sacrificing his sons to the Others. \\
 &  & Olenna & Olenna Tyrell, known as the sharp-tongued and cunning "Queen of Thorns," is the formidable matriarch of House Tyrell. \\
 \midrule
\multirow{22}*{\rotatebox{90}{Avatar: The Last Airbender}} & \multirow{8}*{\rotatebox{90}{main}} & Aang & Aang is the fun-loving, peace-seeking last Airbender and reluctant Avatar tasked with restoring balance to a war-torn world. \\
 &  & Katara & Katara is a compassionate and determined waterbender from the Southern Water Tribe who plays a crucial role in the fight against the Fire Nation alongside Aang and her friends. \\
 &  & Sokka & Sokka is a witty and resourceful warrior from the Southern Water Tribe known for his boomerang skills, inventive mind, and comedic personality. \\
 &  & Zuko & Zuko is the conflicted and determined exiled prince of the Fire Nation whose journey is defined by his quest for honor and self-discovery. \\
\cmidrule(lr){2-4}
 & \multirow{14}*{\rotatebox{90}{minor}} & Iroh & Iroh is a wise and compassionate retired general of the Fire Nation, known for his love of tea, profound spiritual insight, and unwavering support for his nephew Zuko. \\
 &  & Zhao & Zhao is an ambitious and ruthless Fire Nation admiral whose relentless pursuit of power and glory makes him a formidable adversary to Aang and his friends. \\
 &  & Jet & Jet is a charismatic and fiercely determined teenage freedom fighter known for leading a group of rebels against the Fire Nation with a morally ambiguous approach to justice. \\
 &  & Yue & Yue is the gentle and compassionate princess of the Northern Water Tribe whose destiny becomes intertwined with the fate of the Moon Spirit. \\
 &  & Suki & Suki is the skilled and courageous leader of the Kyoshi Warriors known for her exceptional combat abilities and unwavering sense of justice. \\
 &  & Appa & Appa is Aang’s loyal flying bison and steadfast companion known for his gentle nature, immense strength, and ability to soar through the skies. \\
 &  & Pakku & Pakku is a master Waterbender from the Northern Water Tribe known for his strict teaching style and deep sense of tradition. \\
\bottomrule
\end{tabular}
}
\vspace{2mm}
\caption{Information of characters in our experiments (2/2).}
\label{tab:character_info_2}
\end{table}

\end{document}